\documentclass[10pt,journal]{IEEEtran}

\usepackage{amsmath,bm}
\usepackage{amssymb}
\usepackage{stfloats}
\usepackage{graphics}
\usepackage{graphicx}
\usepackage{picins}
\usepackage{color, soul}
\usepackage{bm}
\usepackage{ifthen}
\usepackage{microtype}
\usepackage{wrapfig}
\usepackage{url}
\usepackage{subfigure}
\usepackage{algorithm}
\usepackage{multirow}
\usepackage{algpseudocode}

\ifCLASSOPTIONcompsoc
  \usepackage[nocompress]{cite}
\else
  \usepackage{cite}
\fi

\ifCLASSINFOpdf
\else
\fi

\begin{document}

\title{Image Retargetability}

\author{Fan~Tang, Weiming Dong~\IEEEmembership{Member,~IEEE,} Yiping~Meng, Chongyang~Ma, Fuzhang~Wu, Xinrui~Li, Tong-Yee~Lee,~\IEEEmembership{Senior Member,~IEEE}
\IEEEcompsocitemizethanks{
\IEEEcompsocthanksitem Fan Tang is with NLPR, Institute of Automation, Chinese Academy of Sciences, Beijing, China and University of Chinese Academy of Sciences, Beijing, China. E-mail: tangfan2013@ia.ac.cn.
\IEEEcompsocthanksitem Weiming Dong is with NLPR, Institute of Automation, Chinese Academy of Sciences, Beijing, China. E-mail: weiming.dong@ia.ac.cn.
\IEEEcompsocthanksitem Yiping Meng is with Didi Chuxing, Beijing, China. E-mail: mengyipingkitty@didichuxing.com
\IEEEcompsocthanksitem Chongyang Ma is with Kuaishou Technology, United States. E-mail: chongyangm@gmail.com
\IEEEcompsocthanksitem Fuzhang Wu is with Institute of Software, Chinese Academy of Sciences, Beijing, China. E-mail: fuzhang@iscas.ac.cn
\IEEEcompsocthanksitem Xinrui Li is with Department of Mathematics and Physics, North China Electric Power University, Beijing, China. E-mail: szyclxr@163.com
\IEEEcompsocthanksitem Tong-Yee Lee is with National Cheng Kung University, Tainan, Taiwan. E-mail: tonylee@mail.ncku.edu.tw}
\thanks{}}

\markboth{IEEE Transactions on Multimedia,~Vol.~XX, No.~XX, 2019}%
{Tang \MakeLowercase{\textit{et al.}}: Image Retargetability}

\IEEEtitleabstractindextext{%
\begin{abstract}

Real-world applications could benefit from the ability to automatically retarget an image to different aspect ratios and resolutions while preserving its visually and semantically important content.
However, not all images can be equally processed.
This study introduces the notion of image retargetability to describe how well a particular image can be handled by content-aware image retargeting.
We propose to learn a deep convolutional neural network to rank photo retargetability, in which the relative ranking of photo retargetability is directly modeled in the loss function.
Our model incorporates the joint learning of meaningful photographic attributes and image content information, which can facilitate the regularization of the complicated retargetability rating problem.
To train and analyze this model, we collect a dataset that contains retargetability scores and meaningful image attributes assigned by six expert raters.
The experiments demonstrate that our unified model can generate retargetability rankings that are highly consistent with human labels.
To further validate our model, we show the applications of image retargetability in retargeting method selection, retargeting method assessment and generating a photo collage.

\end{abstract}

}

\maketitle

\IEEEdisplaynontitleabstractindextext

\IEEEpeerreviewmaketitle

\section{Introduction}

\IEEEPARstart{C}{ontent-aware} image retargeting (CAIR) addresses the increasing demand of display images on devices with varying resolutions and aspect ratios while preserving its visually important content and avoiding observable artifacts~\cite{Avidan:07,Rubinstein:09,Pritch:09,Panozzo:2012,Lin:2013:PBI,Sun:2013:SOA,Zhang:2015:RSR}.
Although state-of-the-art image retargeting techniques can successfully handle numerous images, whether a specific image can be successfully retargeted beforehand remains unclear.
CAIR techniques typically expect that the input image contains a mid-sized salient object and a relatively simple background, for which the majority of information can be presented in a small space.
The retargeting results may present severe artifacts if the input images contain rich contents or geometric structures that may be damaged.
Furthermore, not all CAIR methods work equally well for the same input.
The optimal approach which considers quality and robustness depends on the input image and target resolution.
For example, warping-based retargeting methods~\cite{Wang:08,Panozzo:2012,Lin:2013:PBI} are effective and popular, but tends to overstretch or oversqueeze some contents when salient shapes should be preserved.

\begin{figure*}
\includegraphics[width=1.0\linewidth]{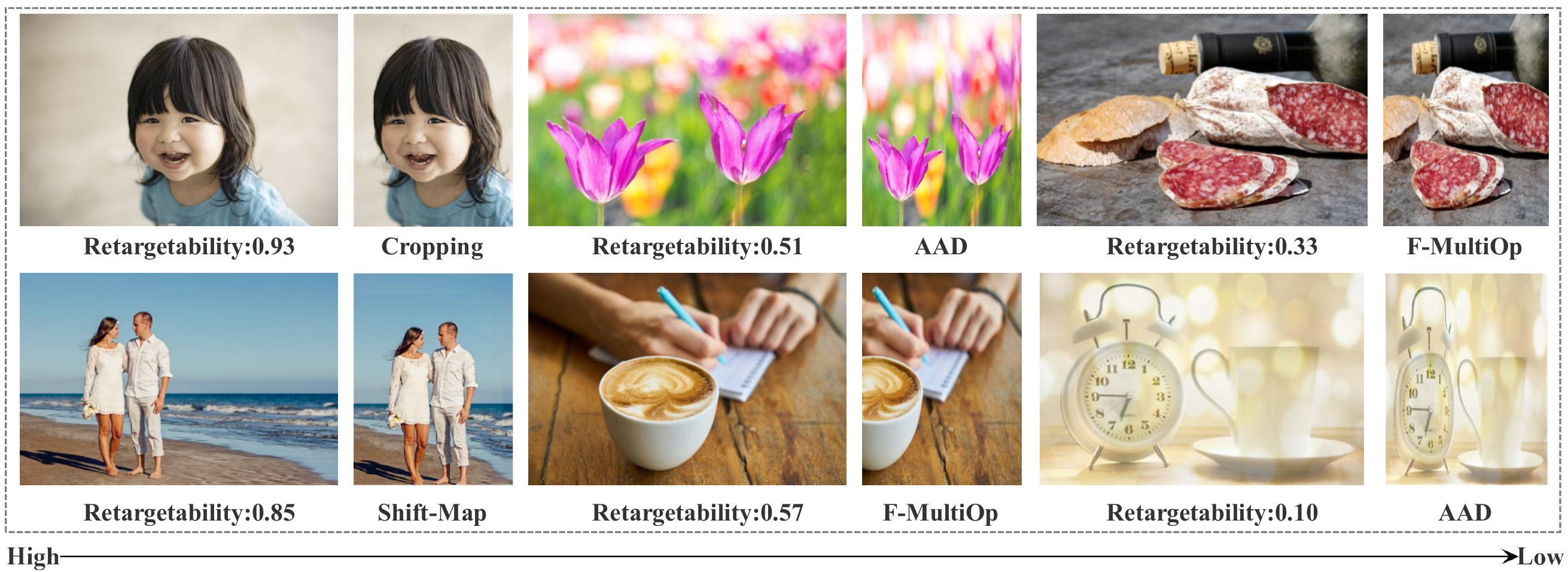}
   \caption{Retargetability of images predicted by our system. The values are in $[0, 1]$, in which a high value indicates that the image is easy to retarget. For each group, left is the original image and right is the retargeted result generated by CAIR method suggested by our system.
}
   \label{fig:teaser}
\end{figure*}

To address the problems of the CAIR method selection and result evaluation, we introduce the notion of ``image retargetability'' to quantitatively compute how well the image can be retargeted on the basis of its visual content.
Fig.~\ref{fig:teaser} shows the predicted retargetability scores of several input images and the corresponding results of the ``best'' retargeting method selected by our system.

We are inspired by some recent studies on quantifying qualitative image properties, such as interestingness~\cite{Gygli:2013:II}, memorability~\cite{Isola:2014:WMP}, synthesizability~\cite{Dai:2014:STE}, and mirrorability~\cite{Yang:2015:MMT}.
To compute image retargetability, we adopt a data-driven methodology and collected a dataset of $13,584$ sample images from Internet photos (Section~\ref{sec:dataset_preparation}).
For each image in the dataset, we apply multiple retargeting methods and request six expert raters to label the quality of each retargeting result in one of the following three levels: good, acceptable, and bad (see Fig.~\ref{fig:website} for examples).
We also ask the raters to annotate a set of high-level visual attributes for each sample image in the dataset, including repeating patterns, specific geometric structures, perspective, fuzzy, text, and shading contrast.

We propose to quantitatively measure and analyze image retargetability on the basis of the collected dataset with manual annotations.
We demonstrate that there is a strong correlation between image retargetability and other visual attributes.
We use this observation as basis to leverage a deep convolutional neural network (NN) and propose a multi-task learning approach by jointly learning visual attributes from deep features and feature sharing for retargetability prediction (Section~\ref{sec:modeling_retargetability}).
We evaluate the effectiveness of our framework for image retargetability prediction by comparing against a baseline approach in Section~\ref{sec:evaluations}.
Given that each CAIR method exhibits its own advantages and limitations, no single CAIR algorithm that works better than other algorithms in all the cases has been produced.
We demonstrate how to select the ``best'' CAIR method using our system.
We also show that image retargetability is useful for retargeting method assessment and photo collage generation (Section~\ref{sec:applications}).

 In summary, our main contributions are as follows.
\begin{itemize}
\item
We introduce image retargetability as a new quantitative property for image analysis.

\item
We collect a large image dataset to learn deep features for image retargetability prediction. The dataset and source code will be released upon final publication.

\item
We adopt a deep NN and propose a novel multi-task learning architecture to compute the retargetability of a given image.

\item
We demonstrate that image retargetability can facilitate several applications for image analysis/processing, such as retargeting method assessment/selection and generating a photo collage.

\end{itemize}

\section{Related Work}

\textbf{Image retargeting algorithms.}
The concept of CAIR aims to preserve the important content of an image after resizing. 
Cropping has been widely used to eliminate unimportant information from the image periphery or improve the overall composition of an image ~\cite{Yan:2013:LCA,Sun:2013:SOA,Zhang:2014:WSP}.
However, cropping often destroys object completeness and causes unexpected information losses.
Discrete methods remove or insert pixels or patches judiciously to preserve content.
Seam carving methods iteratively remove a seam in the input image to preserve visually salient content~\cite{Avidan:07,Rubinstein:08}.
Shift-map method~\cite{Pritch:09} performs a discrete labeling over individual pixels and retargets an image by removing segments in the net.
These approaches are good at retargeting images with rich texture content but may occasionally cause local discontinuity artifacts.
Continuous methods focus on preserving local structure and often optimize a warping from the source size to the target size, constrained on its important regions and permissible deformation~\cite{Wolf:07,Wang:08,Zhang:08,Krahenbuhl:2009}.
Panozzo et al.~\cite{Panozzo:2012} minimized warping energy in the space of axis-aligned deformations (AAD) to avoid unnatural distortions.
Kaufmann et al.~\cite{Kaufmann:2013:FEI} adopt a finite element method to formulate image warping.
Lin et al.~\cite{Lin:2013:PBI} presented a patch-based scheme with an extended significance measurement to preserve the shapes of visually salient objects and structural lines.
Tan et al.~\cite{Tan:2016:IRP} generated feature-preserving constraints in the space of AAD by calculating a feature salience map to guide the warping process.
These approaches can smoothly preserve the image content geometric structure but may also permit minimally important and unwanted regions to appear in the retargeting result. 
Multi-operator methods~\cite{Rubinstein:09,Dong:2009c,Dong:2012} fuse three condensation operators (i.e., seam caring, cropping, and scaling) into a unified optimization framework.
Different operators influence one another and are simultaneously optimized to retarget images.
Summarization-based methods measure patch similarity and select patch arrangements that fit well together to change image size~\cite{Simakov:08,Barnes:09,Dong:2016:IRT}.

Researchers recently adopted deep learning techniques to solve CAIR and related tasks~\cite{Chen:2017:LCP,Wang:2017:DCA,Debang:2018:A2,Chen:2018:CRT}.
Guo et al.~\cite{Guo:2017:AIC} cropped aesthetically pleasing regions on the basis of a novel cascaded cropping regression method.
Song et al.~\cite{Song:2018:PSD} proposed a two-module deep architecture to encode the human perception for image retargeting task and perform a multi-operator-based photo squarization solution.
These deep learning-based approaches are the extension of traditional cropping, warping, or multi-operator-based methods. 
Section~\ref{sec:dataset_methods} provides the details of the advantages and disadvantages of the different types of CAIR methods.

\textbf{Image retargeting evaluations.}
Rubinstein et al.~\cite{Rubinstein:2010} present the first comprehensive perceptual study and analysis of image retargeting, created the ¡°RetargetMe¡± benchmark, and conducted a user study to compare the retargeted images generated by numerous state-of-the-art methods.
An overall ranking of the retargeting methods has been provided on the basis of user study.
Liu et al.~\cite{liu2011image} proposed an objective quality assessment metric that simulates the human vision system to compare image quality with different retargeting methods. 
Their experiments also suggest that no single method is absolutely superior to others in all the cases.
Ma et al.~\cite{Ma:2012:IRQ} built an image retargeting quality data set to analyze different retargeting factors, including scales, methods, and image content.
Zhang et al.~\cite{Zhang:2014:OQE} analyzed three determining factors for the human visual quality of experience, namely, global structural distortion, local region distortion, and loss of salient information.
Fang et al.~\cite{Fang:2014:OQA} generated a structural similarity map to evaluate if the structural information is well preserved in the retargeted image.
Hsu et al.~\cite{Hsu:2014:Objective} proposed a novel full-reference objective metric for assessing the visual quality of a retargeted image on the basis of perceptual geometric distortion and information loss.
Bare et al.~\cite{Bare2014Learning} proposed a new feature and predicted the retargeted image quality by training an RBF NN.
Wang et al.~\cite{Wang:2015:where2stand} analyzed human-scenery position relationship, which can be used to evaluate content composition, in retargeted images.
Zhang et al.~\cite{Zhang2016Backward} adopted a novel aspect ratio similarity metric to measure the geometric change of the images as proven by how the original image is retargeted.
Liang et al.~\cite{Liang:2017:OQP} evaluated image retargeting quality through multiple factors, including preservation of salient regions, symmetry, and global structure, influence of artifacts, and aesthetics. 
Eye tracking data are also used to improve the performance of the objective quality metrics for retargeted image~\cite{castillo2011using}.
Rawat et al.~\cite{Rawat:2018:Spring} focused on the visual balance of social media images and provide real-time feedback on the relative size of image frame.
Several studies~\cite{Hsu:2014:Objective,Bare2014Learning} have also learned to predict a score to discover how well the retargeted images are to indicate whether the quality of a specific retargeting result is good.
However, these studies evaluated the image retargeting quality by comparing the original and retargeted images. By contrast, the current study focuses on predicting the quality of retargeting result from the input image itself, thereby possibly indicating if an image can be well retargeted.

\textbf{Image property analysis.}
Various semantic properties of images have been widely analyzed. 
Rosenholtz et al.~\cite{rosenholtz2007measuring} measured the visual clutter of an image, which is useful for the retrieval of visual content.
Recently, unusual photographs are found to be interesting~\cite{Gygli:2013:II}, and images of indoor scenes with people are found to be memorable, whereas scenic and outdoor scenes are not~\cite{Isola:2014:WMP}.
Other qualitative image properties such as popularity~\cite{khosla2014makes}, colorfulness~\cite{Amati:2014:SIC}, and aesthetics~\cite{Lu:2015:RIA} have been also studied.
Dai et al.~\cite{Dai:2014:STE} used the techniques of example-based texture synthesis as bases to quantify texture synthesizability as an image property, which can be learned and predicted.
In text-based image retrieval, image specificity~\cite{Jas:2015:IS} based on image content and properties has been used to discriminate easily describable images.

The current research defines image retargetability as a semantic property to quantify the probability that an image can be well retargeted.
We show that this notion is closely related to deep relative attributes~\cite{Yang:2016:DRA}.
\section{Dataset Preparation}
\label{sec:dataset_preparation}

This section introduces our data set preparation for image retargetability investigation.
First, we collect a large set of images and manually label each image with a few attributes on the basis of visual content (Section~\ref{sec:img_attri}).
Second, we apply four typical CAIR methods to all the images in the dataset and manually annotate the quality of each retargeting result (Section~\ref{sec:dataset_methods}).

\subsection{Images and Attributes}
\label{sec:img_attri}

Our framework is designed to measure image retargetability on a wild spectrum of natural images.
Accordingly, the dataset should be considerable variability in terms of contents and compositions.
Although the ``RetargetMe'' benchmark~\cite{Rubinstein:2010} has been widely used in image retargeting works for quality assessment, this dataset only contains $80$ images, which are inadequate for the reliable learning of image attributes.
To learn retargetability prediction, we prepare an image dataset and manually annotate the sample inputs in terms of retargetability.
We collect $14,000$ images from Flickr, Pinterest, 500px, and Pexels under Creative Commons license by providing $26$ keywords acquired from 500px photo categories (see \url{https://500px.com}).
The keywords cover the most common categories, such as animals, food, nature, sport, travel, still-life, fashion, and urban exploration. All images are homogeneously scaled by truncating their long sides to $500$ pixels.
Images smaller than this size are not used.
We remove some images that are of low quality or heavily watermarked.
Lastly, we add the ``RetargetMe'' images and ended up with a data set of $13,584$ images.

\begin{figure}
    \centering
    \includegraphics[width=\linewidth]{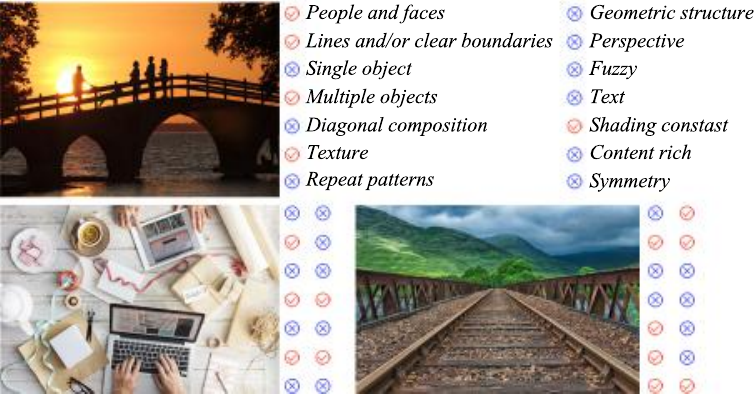}
    \caption{Example images in our dataset with manually annotated attributes.}
    \label{fig:attributes}
    \vskip -0.5cm 
\end{figure}

\begin{figure}
    \centering
    \includegraphics[width=0.95\linewidth]{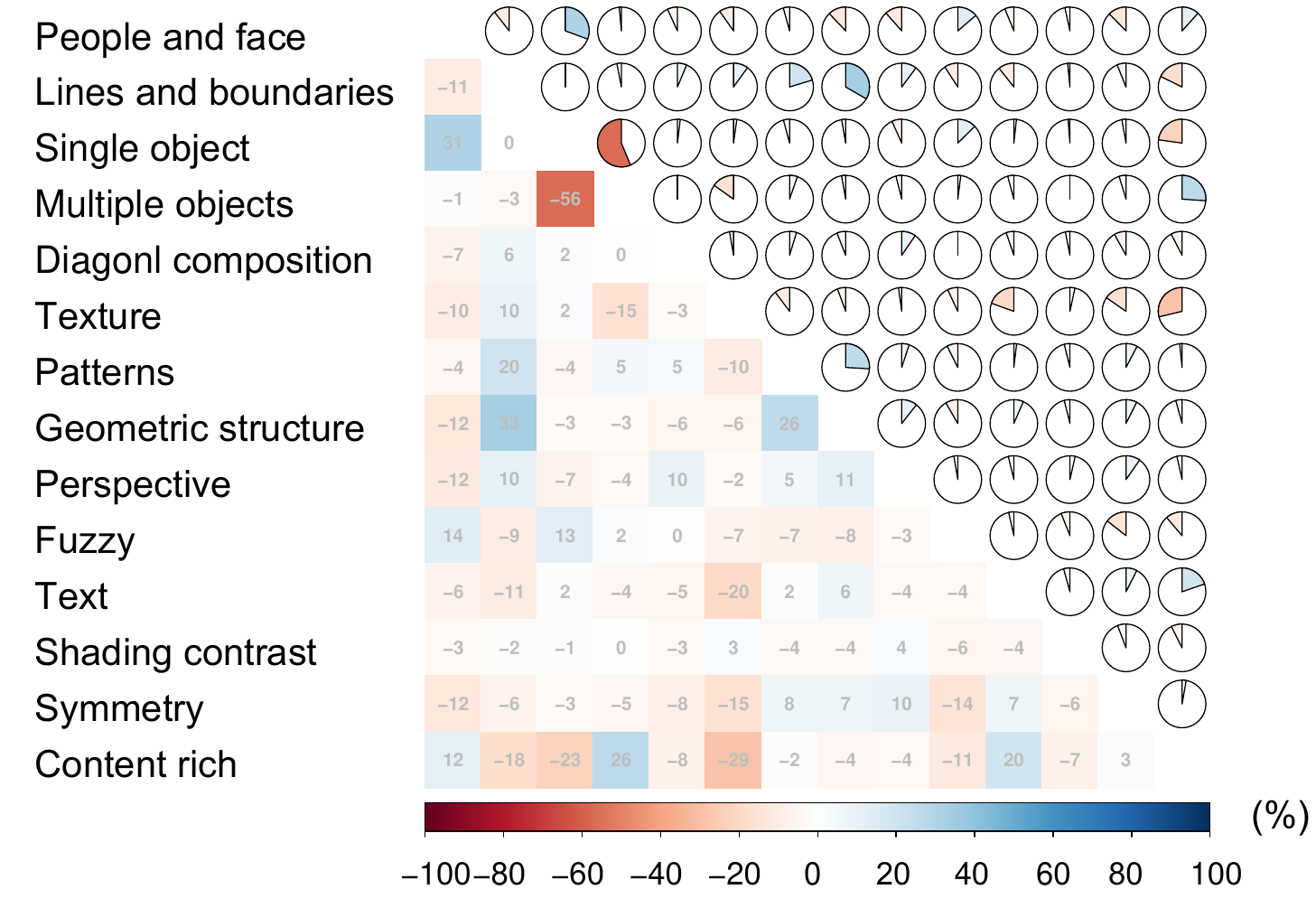}
    \caption{Correlation among the visual attributes.}
    \label{fig:statistics}
    \vskip -0.5cm 
\end{figure}

The CAIR methods work best on images with disposable content.
These images typically include either smooth or some regularly textured areas, such as sky, water, or grass.
Challenges are present when the input image contained either rich semantic contents, salient texts, or geometric structures that may be damaged during retargeting.
We use this observation and photography theories~\cite{Yao:2012:OOS} as bases to choose a set of attributes that can be mapped to the several major retargeting objectives (preserving content, structure, and aesthetics and preventing  artifacts) and manually annotate collected images with these attributes.
The selected attributes are \textit{people and faces}, \textit{lines and/or clear boundaries}, salient \textit{single object}, salient \textit{multiple objects}, \textit{diagonal composition}, \textit{texture}, \textit{repeating patterns}, specific \textit{geometric structures}, \textit{perspective}, \textit{fuzzy}, \textit{text}, \textit{shading contrast}, \textit{content rich}, and \textit{symmetry}.
Fig.~\ref{fig:attributes}  shows some examples in our data set with the attributes assigned to each image.
Fig.~\ref{fig:statistics} shows the correlation between image attributes.
Apart from some attributes with opposite meaning (e.g., a single object versus multiple objects), the majority of the numbers in Fig.~\ref{fig:statistics} contain relatively low absolute values, thereby demonstrating that most attributes are uncorrelated to one another.

\subsection{Retargeting Methods and Annotations}
\label{sec:dataset_methods}

To evaluate the retargetability of the collected images in the dataset, we select and implement the four most typical and commonly used CAIR methods, namely, multi-operators, inhomogeneous warping, shift-map, and cropping.
We apply these four methods to all the collected images in the dataset.

\begin{itemize}
\item
\emph{Multi-operator} method outperforms the majority of the other approaches according to the comparative study~\cite{Rubinstein:2010}.
A typical multi-operator method integrates seam carving, homogeneous scaling, and cropping to resize an image and can be considered a generalized version of seam carving. Our study adopts the fast multi-operator method~\cite{Dong:2012}, which is sufficiently rapid for practical applications.

\item
\emph{Inhomogeneous warping}-based method is known for its real-time performance and local continuity preservation.
We use the AAD method~\cite{Panozzo:2012}, which has been recently verified to be one of the most effective warping methods.
Other state-of-the-art warping-based methods~\cite{Krahenbuhl:2009,Lin:2013:PBI,Kaufmann:2013:FEI} can also be used as the representative method, which does not affect the effectiveness of our retargetability learning and prediction framework.

\item
\emph{Shift-map}-based method can selectively stitch some contents together and often works well for input with salient contents distributed in the different parts of the image.
Our study applies the original shift-map method~\cite{Pritch:09} to the images in the data set.

\item
\emph{Cropping}-based CAIR algorithm is preferred in many cases because this type of method does not introduce any distortion in the retargeting results~\cite{Rubinstein:2010}.
In particular, we use the SOAT$_\text{\rm cr}$ method~\cite{Sun:2013:SOA}.
\end{itemize}

\textbf{Discussions.}
We did not add summarization-based retargeting methods, such as BDS~\cite{Simakov:08} and PM~\cite{Barnes:09}, during data collection for two reasons. This method typically requires several minutes to generate a good result, while the results of BDS/PM often present a structural mismatch of spatial content~\cite{Dong:2016:IRT}.
These two artifacts limit their practical use in many applications, particularly for some systems that require real-time performance.
We did not integrate some other CAIR methods that focus on images containing specific contents, such as symmetry structure~\cite{Wu:2010:RSS}, semantically-rich information~\cite{Zhang:2015:RSR}, and textures~\cite{Dong:2016:IRT}, because our goal was to evaluate image retargetability in a generic manner.
\begin{figure}[!h]
\centering
\vskip -2mm
\includegraphics[width=0.45\linewidth]{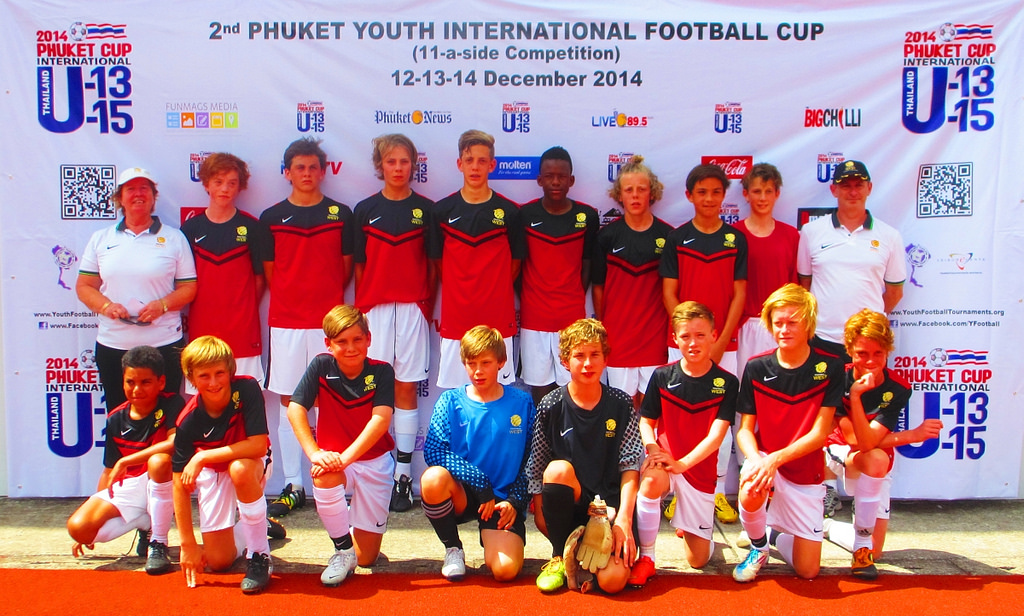}
\hspace{4mm}
\includegraphics[width=0.45\linewidth]{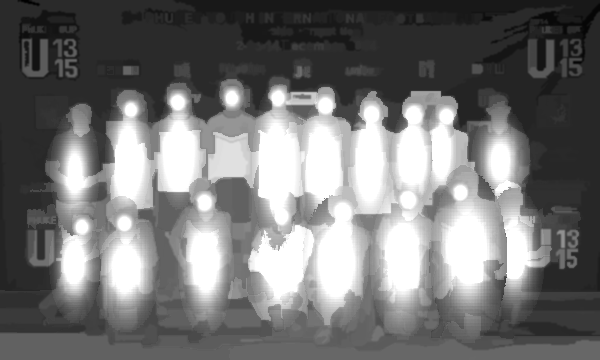}
\caption{Example of the importance map used in our framework.}
\label{fig:importance}
\vskip -2mm
\end{figure}

\begin{figure}[h]
    \centering
    \includegraphics[width=\linewidth]{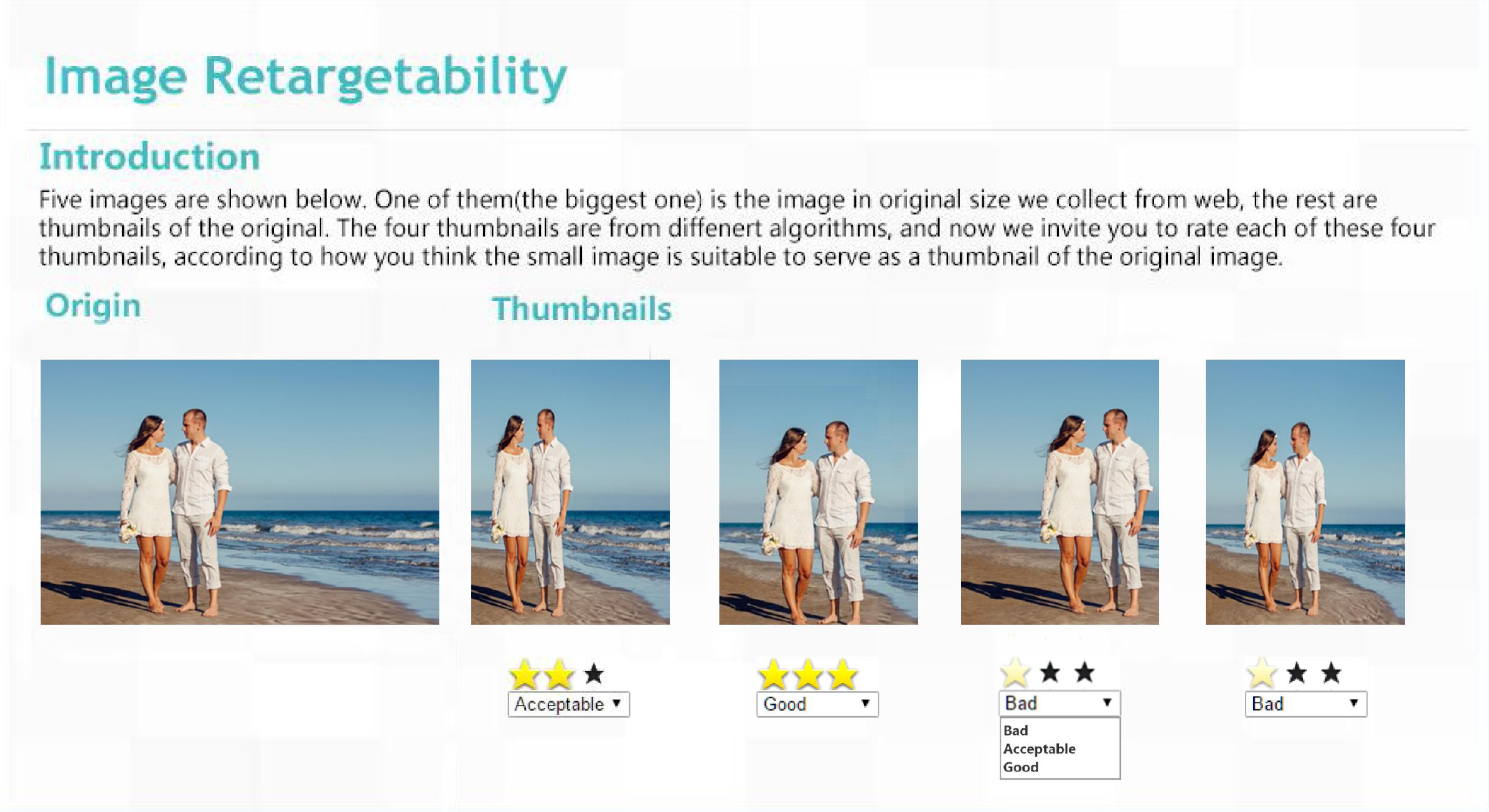}
    \vspace{-2mm}
    \caption{Our web interface for data annotation. Expects were invited to rate each of the retargeted images on the basis of their own opinion. No further information on the definition of retargetability is released.}
    \vspace{-2mm}
    \label{fig:website}
\end{figure}

\begin{figure*}
\centering
\includegraphics[width=0.98\linewidth]{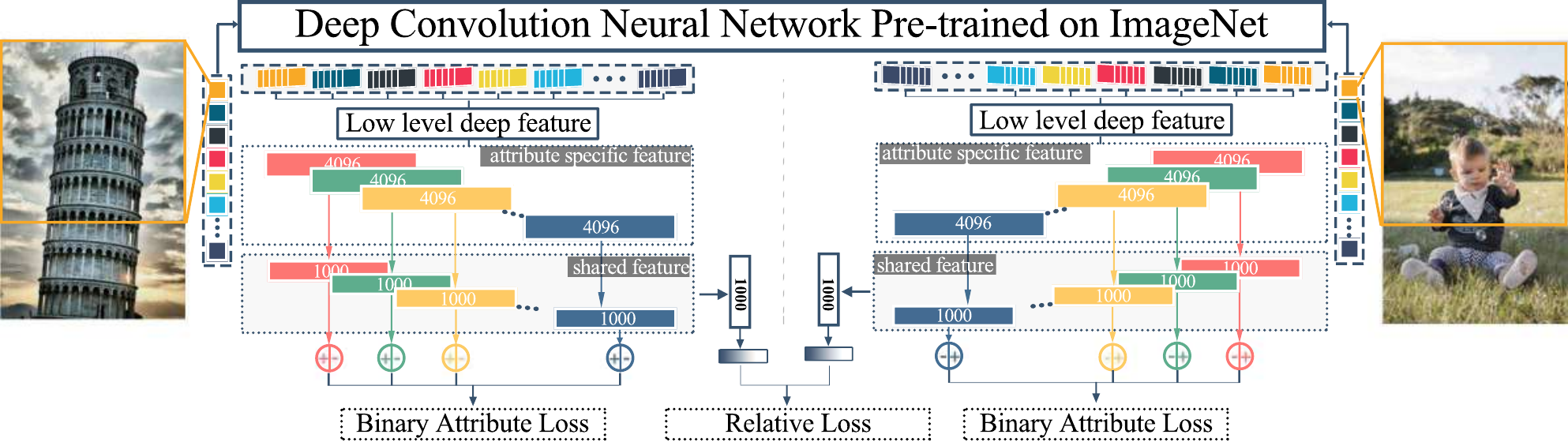}
\caption{Overall structure of our method. A siamese network with a three-level feature representation mechanism and two types of loss function corresponding to binary visual features or relative retargetability was adopted for retargetability learning. }
\label{pipeline}
\vskip -5mm
\end{figure*}

Although the selected methods are not recently proposed, they represent the mainstream CAIR framework.
The recently proposed deep learning-based methods are an extension of these methods.
We choose these classic approaches because they have been widely tested and proven to be stable and effective.

Given that the majority of the CAIR methods are carefully designed for one-dimensional retargeting, we restrict the change to either the width or height of an image.
For each image in the data set, we resized the long dimension to $50\%$ using the four CAIR methods described in Section~\ref{sec:dataset_methods}, all with fixed parameters.
In particular, we choose to retarget the images to half their size, which is similar to the methods performed in previous research on CAIR, because the majority of the images can be handled well for small-sized changes, while causing poor results for large changes.
We further guided the CAIR methods by computing an importance map for each image.
We adopt state-of-the-art saliency detection approaches~\cite{Cheng2015Global}, face segmentation~\cite{Saito2016Real} and body detector~\cite{Ren2015Faster} to generate the importance map.
Note that the output of a body detector is the bounding box of the body region.
We use GrabCut~\cite{rother2004grabcut} to generate the importance map when a body is detected.
We use the average of these maps as our importance map (see Fig.~\ref{fig:importance}).

We ask six expert raters to independently evaluate the quality of all the retargeted images and annotate the result as one of the following three levels (see Fig.~\ref{fig:website}): \emph{good}, \emph{acceptable}, and \emph{poor}, which correspond to scores of $1$, $0.5$ and, $0$, respectively.
Thereafter, we compute the average score from the six raters as the evaluation of each retargeting result in the dataset.

\textbf{Consistency Analysis of Annotations}
We measure the inter-rater consistency to verify the objectivity of the annotation data.
For each image in our data set, we adopt Kendall's coefficient of concordance (Kendall's W)~\cite{kendall1939}  to study the rating consistency among different subjects.
Kendall's W is a non-parametric statistical measure that ranges from $0$ (no consistency) to $1$ (completely consistent).
The overall average Kendall's W is $0.562$ with a standard deviation of $0.0192$.
Moreover, raters can obtain significant concordance on $87.69\%$ images at the $0.05$ significance level.

\section{Modeling Retargetability}
\label{sec:modeling_retargetability}

\begin{figure}
    \centering
    \includegraphics[width=0.95\linewidth]{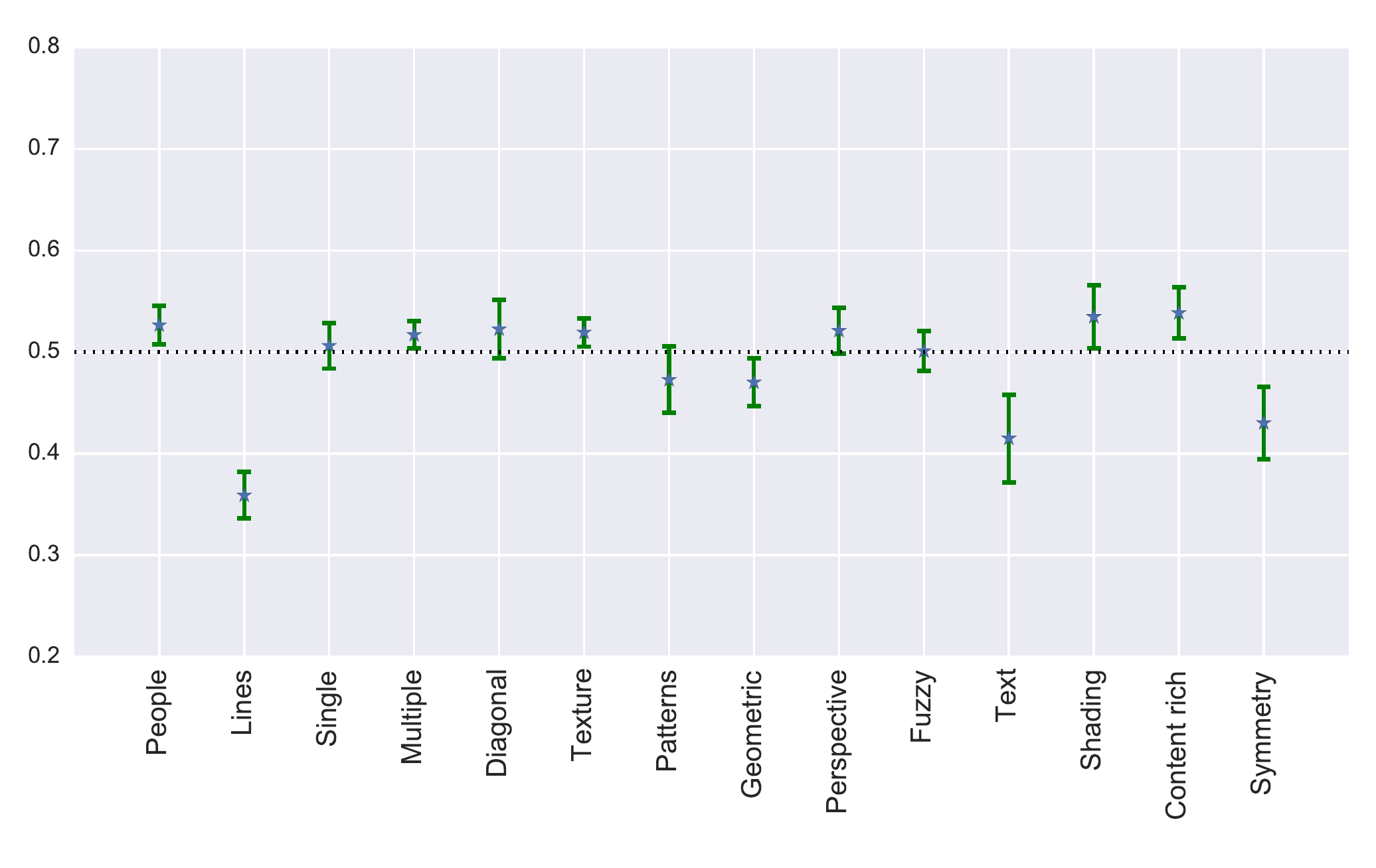}
    \vspace{-3mm}
    \caption{Ridit scores with $95\%$ confidence interval for visual attributes.}
    \label{ridit}
    \vskip -0.5cm 
\end{figure}

First, we use our dataset as basis to analyze the correlation between image attributes and retargetability (Section~\ref{sec:measuring_retargetability}).
Second, we introduce our framework to employ deep learning and multi-task learning to learn and predict image retargetability (Section~\ref{sec:learning}).

\subsection{Measuring Retargetability}
\label{sec:measure}
\label{sec:measuring_retargetability}

For each sample image in the dataset, we define its retargetability as the max score of the four average user-rated scores (each image possesses four retargeted outputs, while each output image contains six user-rated scores; see Section~\ref{sec:dataset_preparation}).

Given this quantitative measurement of retargetability, we can analyze the relationship between different visual attributes and the image retargetability using Ridit analysis~\cite{devore2015probability}, which is commonly used in the study of ordered categorical data.
In Fig.~\ref{ridit}, the dashed horizontal line is the reference unit $0.5$ and the deviation from reference unit represents the influence of the attribute.
Evidently, some visual attributes are closely related to image retargetability.
For example, the groups of lines, text, symmetry, geometry, and patterns are under the dashed line, thereby indicating that images with these attributes are likely to exhibit low retargetability scores or equivalently, worse retargeting results annotated by our raters.
Images with content rich, diagonal structure, and texture often correspond to high scores.
We use this key observation as basis to propose to learn and predict retargetability on the basis of visual attributes.

\subsection{Learning and Predicting Retargetability}
\label{sec:learning}
Although the retargetability of an image is calculated by the ratings of the retargeted images, retargetability itself is a high-level property of such an image.
Thus, we aim to learn retargetability directly from the source image rather than the retargeted images by utilizing the pre-selected attributes to regularize a pair-wise retargetability ranking training.
Fig.~\ref{pipeline} shows the overall structure of our model, including a three-level feature representation mechanism and two types of loss functions that correspond to binary visual features or relative retargetability.
First, we use the output from deep convolutional network as the low-level representation of the image.
Second, we learn attribute-specific features for each attribute and eventually use this information to learn retargetability.
To boost training phase, we simultaneously learn the visual attribute features with retargetability.

In the following section, we demonstrate the multi-task learning approach by jointly learning visual attributes from deep features and feature sharing for retargetability.

\subsubsection{Deep Features}
We use a VGG-19~\cite{Simonyan2014Very} style model pre-trained on ImageNet~\cite{ILSVRC15} for image classification to extract deep representations for input images.
The network consists of a stack of convolution layers with pooling and ReLU, followed by three fully-connected layers and softmax with loss.
After isotropically re-scaling the input image's short side to $224$, we densely crop the image to obtain $224^2$ sub-images and feed the square sub-images to the convolutional network as~\cite{Simonyan2014Very}.
The need to re-scale or crop the image to the same size for the learning appears to defeat the purpose of studying the effect of changing the aspect ratio of an image.
However, compared with image retargetability, other visual attributes are more robust to the size change of the input image.
That is, if an image contains a face, then the face will continue to exist even if the aspect ratio of the image has been changed.
We denote $Fm_i$ as the last convolutional layer's output of $i_{th}$ sub-image.
The low-level deep feature of an input is as follows:
\begin{displaymath}
Fm =\frac{\sum_{i=1}^{K}Fm_{i}}{K},
\end{displaymath}
where $K$ is the number of sub-images and is set to $10$ in our implementation.
We use the output of the convolution layers instead of the fully connected layers to acquire the low-level image representation.
Hence, the output is not considerably related to the pre-trained classification task.

\subsubsection{Learning Retargetability}

We attempt to learn the middle-level features for visual attributes and share these features for retargetability to boost learning performance.
In our task, all  visual attributes are labeled as $1$ or $-1$, in which binary attributes are often learned using the classification method.
We formulate retargetability as a type of relative attribute that is powerful in uniquely identifying an image and offer a semantically meaningful method to describe and compare images in the wild~\cite{Yang:2016:DRA}.
We design different losses for different types of attributes.
Given a low-level deep feature space together with annotated attributes and retargetability-labeled image data, we learn a shared attribute-level feature representation by optimizing a joint loss function that favors a pair-wise relative loss and squared hinge loss function with sparsity patterns across binary attributes.

\textbf{Binary attribute features learning.}
Given $M$ semantic attributes, the goal is to learn $M$ binary classifiers jointly.
Each binary classifier is a four-layer NN with one input layer and two hidden layers of $4096$ and $1000$ nodes, respectively, and a one-node output layer, followed by squared hinge loss function.
Inspired by \cite{Liu2009Multi,Abdulnabi2015Multi}, we utilize $l_{2,1}$-norm minimization to boost feature sharing among the different attributes.
Multi-task feature learning via $l_{2,1}$-norm regularization has been studied in many approaches and encourages multiple predictors from different tasks to share similar parameter sparsity patterns.
Given an image $i$, with a $M$-dimension label vector $L_i$ element that is only {1} or {-1}, we proposed the following equation by supposing that the parameters between the two hidden layers for the $k_{th}$ attribute learning MLP server as $w_{k}$:
\begin{displaymath}
loss_{binary}(i) = \sum_{k=1}^{M} \frac{1}{2}[\max (0,1-L_{ik} \cdot L^*_{ik})]^2 + \frac{1}{2}\alpha\| W \|_{2,1},
\end{displaymath}
where $W=[w_{k}]$ and $L^*_{ik}$ is the output of the $k_{th}$ MLP for image $i$ and $\| W \|_{2,1} = \sum_{}^{}\| w_k \| $ is the $l_{2,1}$-norm of the matrix $W$.
We apply $l_{2,1}$-norm to $W$, thereby indicating that the outputs of the first hidden layer in different MLPs are relatively independent (see ``attribute specific feature'' in Fig.~\ref{pipeline}).
By contrast, the outputs of the last hidden layer are boosted by multi-task feature learning technique (see ``shared feature'' in Fig.~\ref{pipeline}).

\textbf{Relative retargetability learning.}
In general, the goal of relative attribute learning is to learn ranking functions for labeled image pairs.
The existing relative attribute learning approaches learn linear functions to map hand-crafted features to relative scores. Inspired by \cite{Hwang2011Sharing}, we collect features learned for each visual attribute as mid-level visual features and use these attribute-related features to train retargetability by a three-layer NN with $1000$ hidden nodes.
All the shared features are concatenated as the input of the three-layer NN.
We define the relative loss as the sum of the contrastive and similar constraints.
Given a pair of images $i$ and $j$ ($i\neq j$), with retargetability $y_i$ and $y_j$ predicted as $y^*_i$ and $y^*_j$, the loss for the image pair ($i, j$) is as follows:
\begin{displaymath}
loss_{relative}(i, j) = I(i, j) \cdot {l}_p(i, j) + (1-I(i, j)) \cdot {l}_q(i, j),
\end{displaymath}
where
\begin{align}
I(i,j) &= \left\{\begin{matrix} \notag
1, & y_i > y _j\\
0, & y_i \sim y_j
\end{matrix}\right.     \\ \notag
{l}_p(i, j) &=  max(0,\tau-(y^*_i-y^*_j)), \\
{l}_q(i, j) &=  \frac{1}{2}(y^*_i-y^*_j)^2, \notag
\end{align}
where $I(i, j)$ is a binary function that indicates whether images $i$ and $j$ exhibit similar retargetability, ${l}_p(i, j)$ denotes the contrastive constraint for ordered image pair ($i, j$), and ${l}_q(i, j)$ donates similar constraint for unordered pairs.
The parameter $\tau$ controls the relative margin among the attribute values when $I(i, j)=1$.

\textbf{Formulations and implementations.}
Given the pair-wise relative loss, we use a two-channel Siamese network as the overall structure~\cite{zagoruyko2015learning}.
Each channel of the network predicts $14$ visual attributes and retargetability together with the attribute-specific features.
The hinge-based binary loss is calculated among each group of the $M$ attributes while relative loss is computed by two predicted retargetability.
The goal of the entire two-channel network is as follows:
\begin{align}
\notag \min_{\Theta}{J_{\Theta}} = & \sum^{i\neq j}_{i,j}{loss_{binary}(i)+loss_{binary}(j) + {loss_{relative}(i,j)} }\\
\notag & + \beta \|\Theta\|_F,
\end{align}
where $\Theta$ stands for all the parameters to be optimized and $\| \Theta\|_F$ is a regression term to penalize overfitting.
Table~\ref{tab:net_con} summarizes the configuration of the proposed method.
Note that the majority of the parameters come from the MLPs for binary attribute learning.
Due to these MLPs are trained in a multi-task learning manner, the parameters are learned ``separately''.
Given $N$ images for training, we could obtain $14 \times N$ independent labels for training these MLPs separately.
Each attribute is trained on one single small model.
Different attributes are further boosted by feature sharing technique.
For calculating the relative loss, we randomly draw a pair of samples from these $N$ images.
Totally we could draw $\mathcal{C}_N^2$ sample pairs for training.
These advantages can possibly train the big network on the proposed data set.
We adopt mini-batch stochastic gradient descent with a batch-size of $64$ and an initial learning rate of $0.01$.
For the full connected layers, we adopt ReLU as the activation function.
In the training stage, we randomly drop out $30\%$ parameters to push the network to learn additional general features.
In the test stage, we use one way of the siamese network to generate the outputs and truncate the predicted value into $[0, 1]$.
\begin{table}[]
\centering
\caption{Net configuration for proposed architecture.}
\vskip -0.2cm 
\label{tab:net_con}
\includegraphics[width=0.95\linewidth]{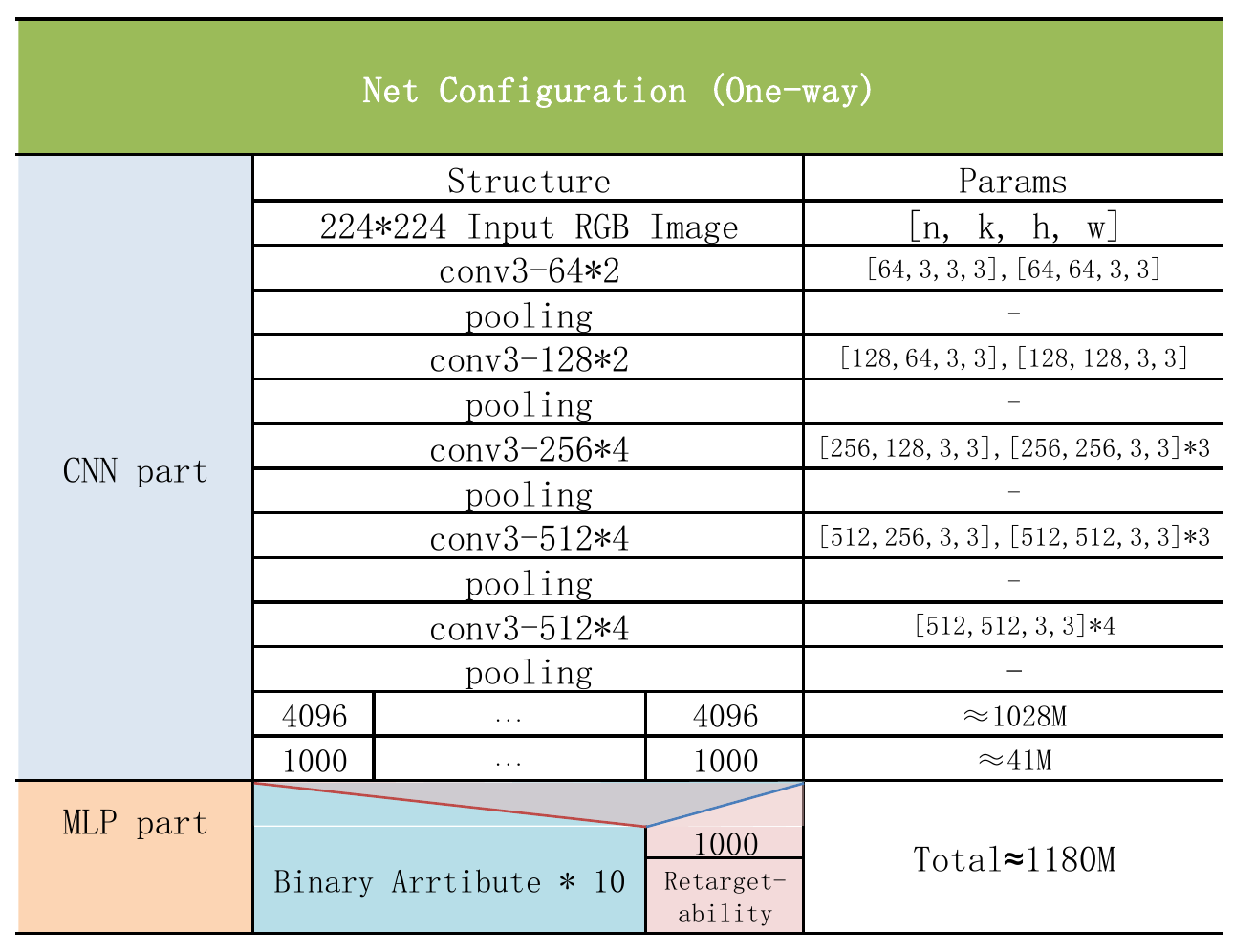}
\vskip -0.5cm 
\end{table}

\section{Evaluations}
\label{sec:evaluations}

This section presents the evaluation and analytical results of the image retargetability prediction.
All experiments are performed on a PC equipped with $3.6$ GHZ Intel Core i7 and Nvidia Geforce GTX $1080$Ti.
The implementation is based on Caffe2.
The proposed network is trained for approximately $6$ hours.
Given the densely cropping operation, the testing speed is approximately $0.3$ fps, which is slower than the current state-of-the-art CAIR method.
Song et al.~\cite{Song:2018:PSD} report that their model takes about $0.5$s to process $100$ images.
However, the current study did not focus on the testing speed.
Accordingly, the speed can be highly improved by taking the model quantization or distillation techniques as~\cite{Song:2018:PSD}.
We randomly select half of the annotated images to train the retargetability predictor and use the remainder for testing.
This process is performed five times and the average results are as follows.
\begin{figure*}[t!]
  \centering
  \includegraphics[width=0.95\linewidth]{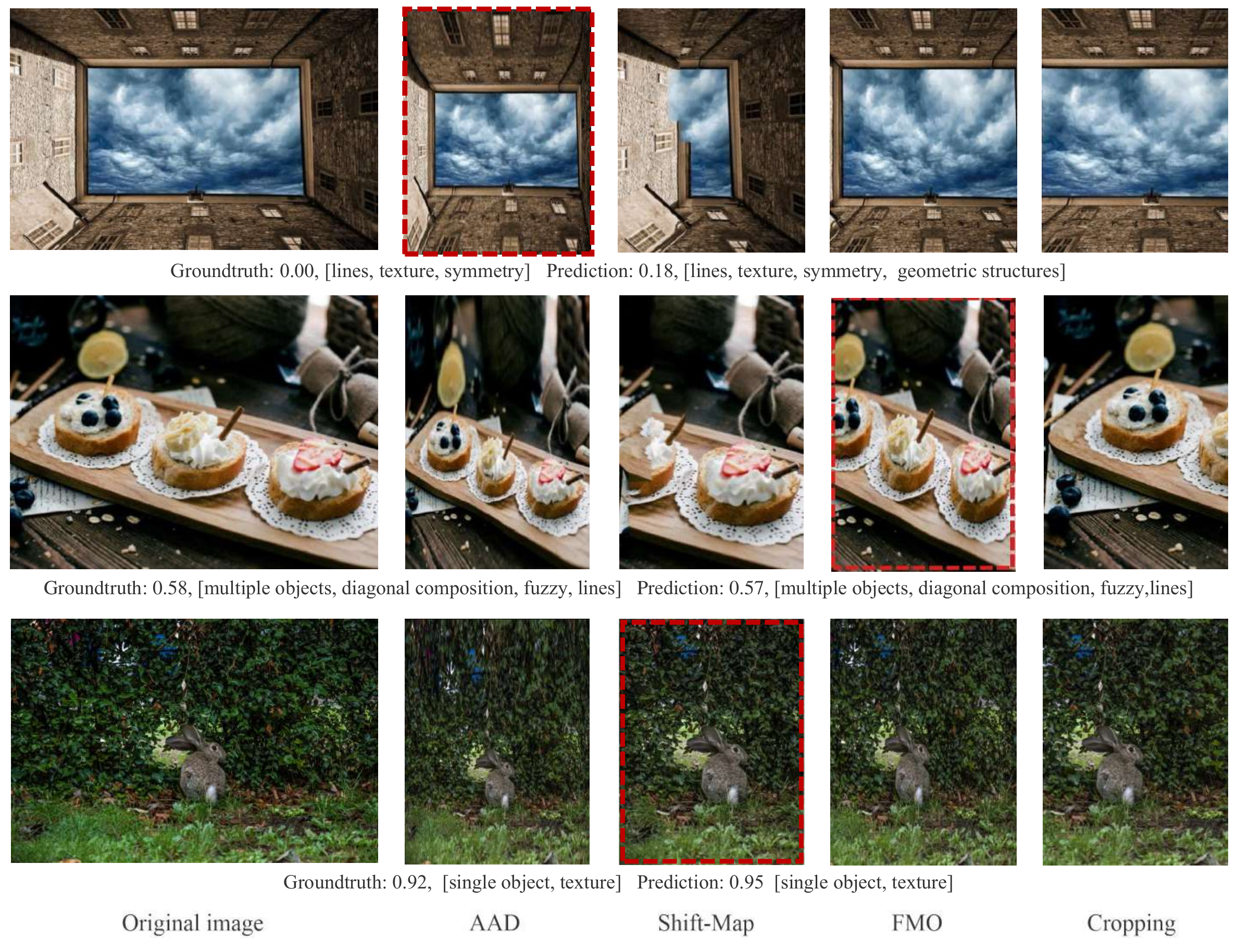}
  \vskip -4mm
  \caption{Examples for retargetability prediction, visual attributes prediction, and method selection. In each example, the input image and the results of the four CAIR methods are shown on the left and right respectively.
  The visual attribute is predicted by the output of MLP for binary attribute learning: if $L^*_k > 0.5$, then the image is labeled as the $k_{th}$ attribute.
  The results of the predicted best method are highlighted with red dash lines.}
  \label{fig:qualitative}
  \vskip -2mm
\end{figure*}

\subsection{Experimental Settings}

Given that our research is the first study of image retargetability, finding direct comparisons with any previous study is difficult.
To demonstrate the effectiveness of our framework, we compare our framework with the following CNN structures:
\begin{itemize}
\item $Net_{-}$. A straightforward end-to-end VGG19, which is fine-tuned on the training data to directly solve the regression problem.
\item $Net_{+}$. A siamese network without binary attribute loss.
All the other configurations are the same as the proposed approach, including the relative loss and low level feature extraction.
\item $Net_{*}$. A siamese network without $l_{2,1}$ normalization in binary attribute loss.
In the proposed method, we utilize $l_{2,1}$ to boost the feature sharing among different binary attributes. $Net_{+}$ is tested to evaluate the performance of $l_{2,1}$ normalization.
\item $Net_{\&}$. A siamese network without dense cropping in low level feature extraction.
We re-scale the input image's short side to $224$ and cropped $K$ sub-images to calculate low-level deep representation.
By contrast, $Net_{\&}$ directly re-scales the input image to $224^2$.
\item $Net_{@}$. A one-way network without relative loss.
All the other configurations are the same as the proposed approach, including the binary attribute loss and low-level feature extraction.
\end{itemize}
Table~\ref{tab} shows the details of these configurations.
\begin{table}[h]
\centering
\caption{Configurations for contrast methods.}
\label{tab}
\begin{tabular}{l|c|c|c|c}
  \hline  \hline
Method      & Dense crop    & Binary attribute  & $l_{2,1}$ loss    & Relative loss \\  \hline
$Net_{-}$   & -             &  -                & -                 & -             \\  \hline
$Net_{+}$   & $\checkmark$  &  -                & -                 & $\checkmark$  \\  \hline
$Net_{*}$   & $\checkmark$  & $\checkmark$      & -                 & $\checkmark$  \\  \hline
$Net_{\&}$  & -             & $\checkmark$      & $\checkmark$      & $\checkmark$  \\  \hline
$Net_{@}$   & $\checkmark$  & $\checkmark$      & $\checkmark$      & -             \\  \hline
$Ours$      & $\checkmark$  & $\checkmark$      & $\checkmark$      & $\checkmark$  \\  \hline  \hline
\end{tabular}
\vskip -0.5cm 
\end{table}

\subsection{Qualitative Analysis}

\begin{figure*}
\centering
 \includegraphics[width=0.99\linewidth]{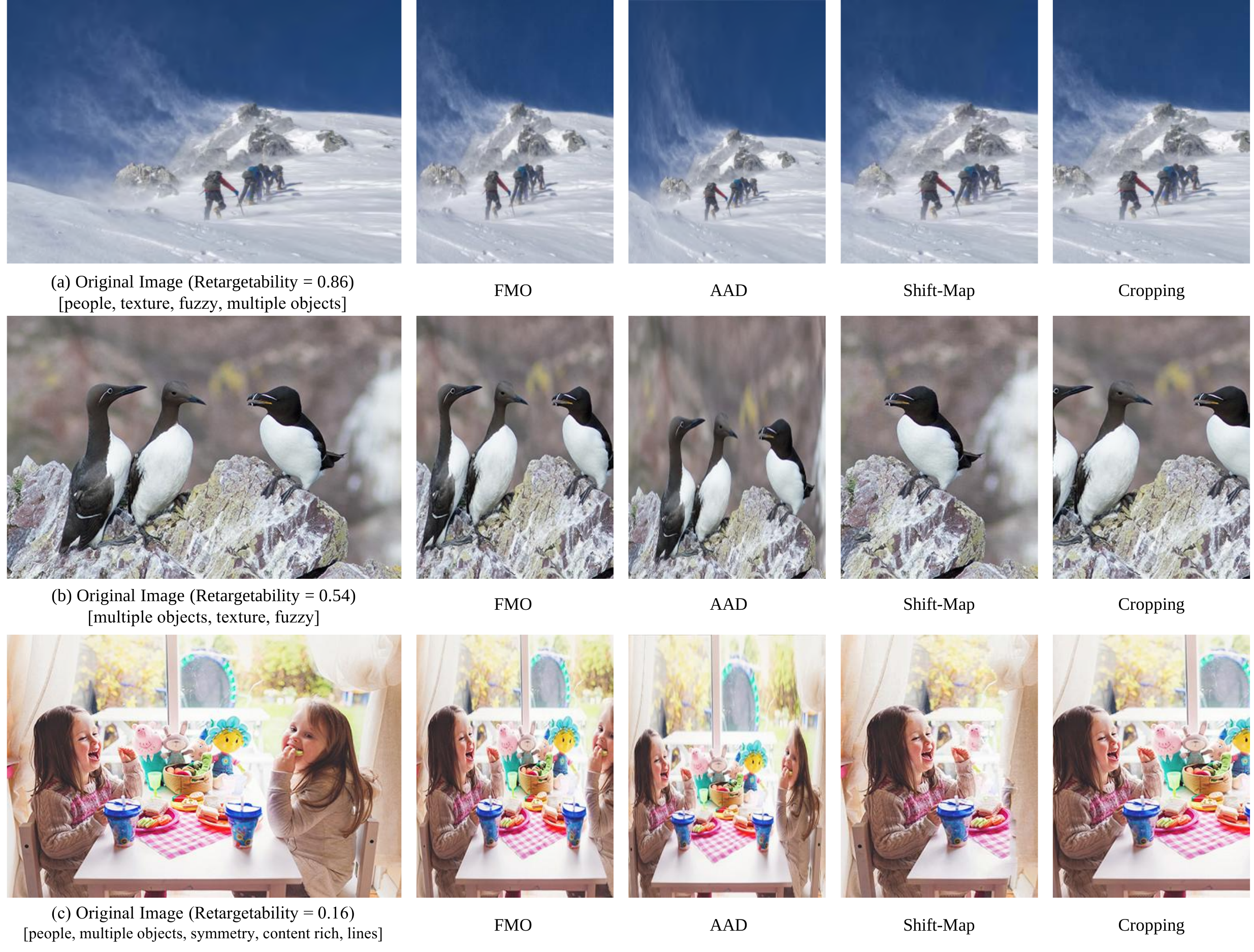}
  \caption{Images with different retargetability scores and the corresponding results from four selected CAIR methods. (b) and (c) are more reliable for assessing new retargeting methods for these images are difficult for existing CAIR methods.}
  \label{fig:app_img_sel}
\end{figure*}

Figs.~\ref{fig:qualitative} and \ref{fig:app_img_sel} show the predicted retargetability and corresponding retargeted output, respectively, for several input images.
The quality of the retargeted images is consistent with the predicted retargetability score.
The results indicate that images with large homogeneous regions, blurry background, and single object lead to increased scores.
By contrast, low scores are caused by several factors, including salient lines, clear boundaries, geometric structures, and symmetry.

\subsection{Quantitative Analysis}

We use root-mean-square error (RMSE) as measurement to evaluate the accuracy of our retargetability prediction approach.
Assuming $N$ images in the testing set are present, the overall RMSE $=\sqrt{\frac{1}{N}\sum_{N}{(y-y^*)^2}}$.
\begin{table}[]
\centering
\caption{RMSE for different net structures.}
\label{tab:RMSE}
\begin{tabular}{llllll|l}
\hline \hline
Method  & $Net_{-}$ & $Net_{+}$ & $Net_{*}$ & $Net_{\&}$    & $Net_{@}$    & Ours   \\ \hline
RMSE    & $0.334$   & $0.296$   & $0.248$   & $0.246$       & $0.228$      & $0.209$ \\ \hline \hline
\end{tabular}
\end{table}
Table~\ref{tab:RMSE} shows the results.
Accordingly, provide the following observation.
\begin{itemize}
\item As a baseline approach, $Net_{-}$ reported the largest RMSE, thereby indicating that the proposed image property and retargetability can not be well-learned using traditional deep convolutional network.
\item The RMSE improvement between $Net_{+}$ and $Net_{*}$ demonstrates the model benefits from the representative ability of the extracted features by joint learning with binary attributes.
\item Dense cropping in low level feature extraction promotes the performance of the feature learning process, which can be proven by the comparisons between $Net_{-}$-$Net_{+}$, $Net_{\&}$-Ours.
Such observations confirm that retargetability is a property dealing with the ability to be resized, retargetable operations to the original images may cause uncertain results.
In our pipeline, retargetability is learned on features related to visual attributes which are considerably insensitive to the input size changing.
\item The proposed model reported the lowest RMSE by embedding all the losses thereby confirming that sharing visual knowledge with high-level image attributes in the predictive model is a compelling method for boosting the learning process.
Compared with dense cropping or binary attribute, the improvement using relative loss is not definite because the proposed relative loss tends to rank the images according to their retargetability.
In training phase, our model is more likely to ``compare'' images with a pair-wise loss rather than learning absolute scores.
\end{itemize}
\begin{figure}
  \centering
  \includegraphics[width=0.99\linewidth]{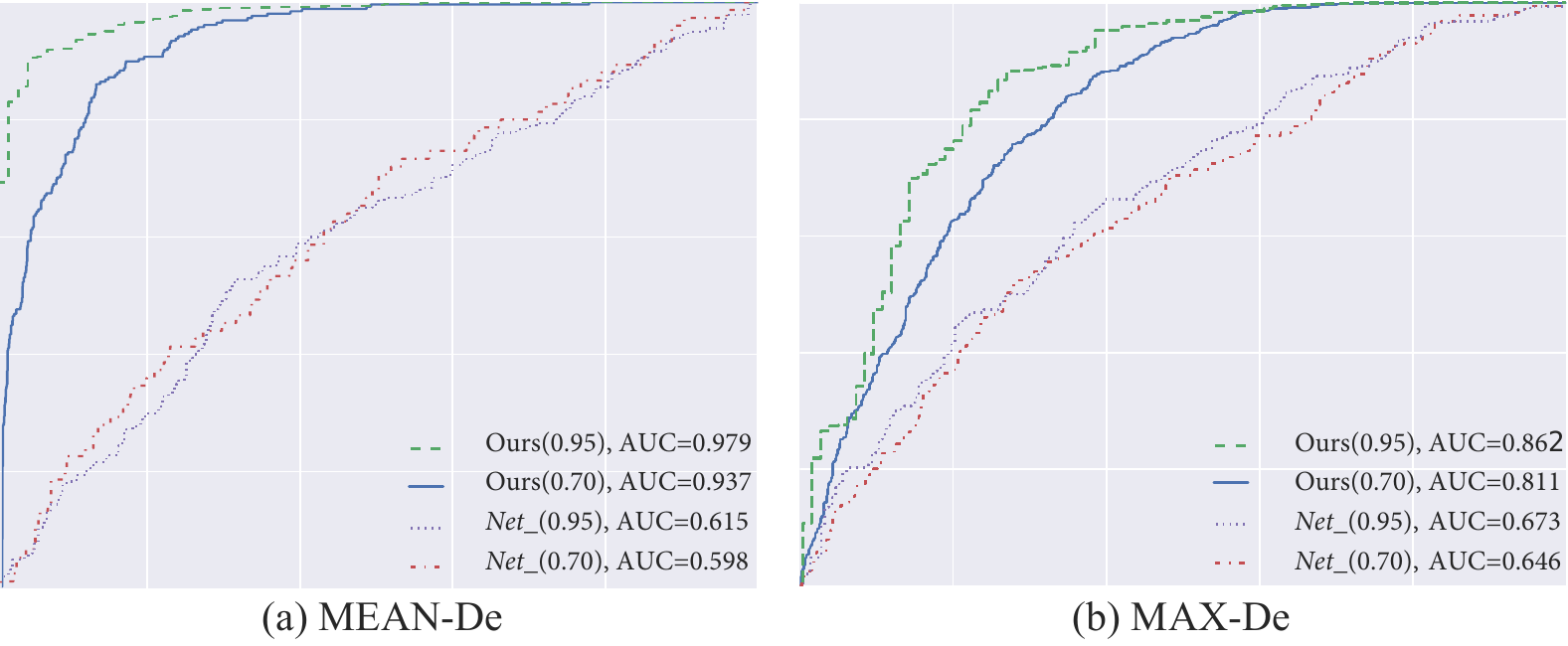}
  \vspace{-4mm}
  \caption{Comparison of the retargetability prediction accuracy between our method and the baseline approach.}
  \label{fig:roc}
\end{figure}

\subsection{Discussion of Definition}

\begin{figure}[h]
  \centering
  \includegraphics[width=0.99\linewidth]{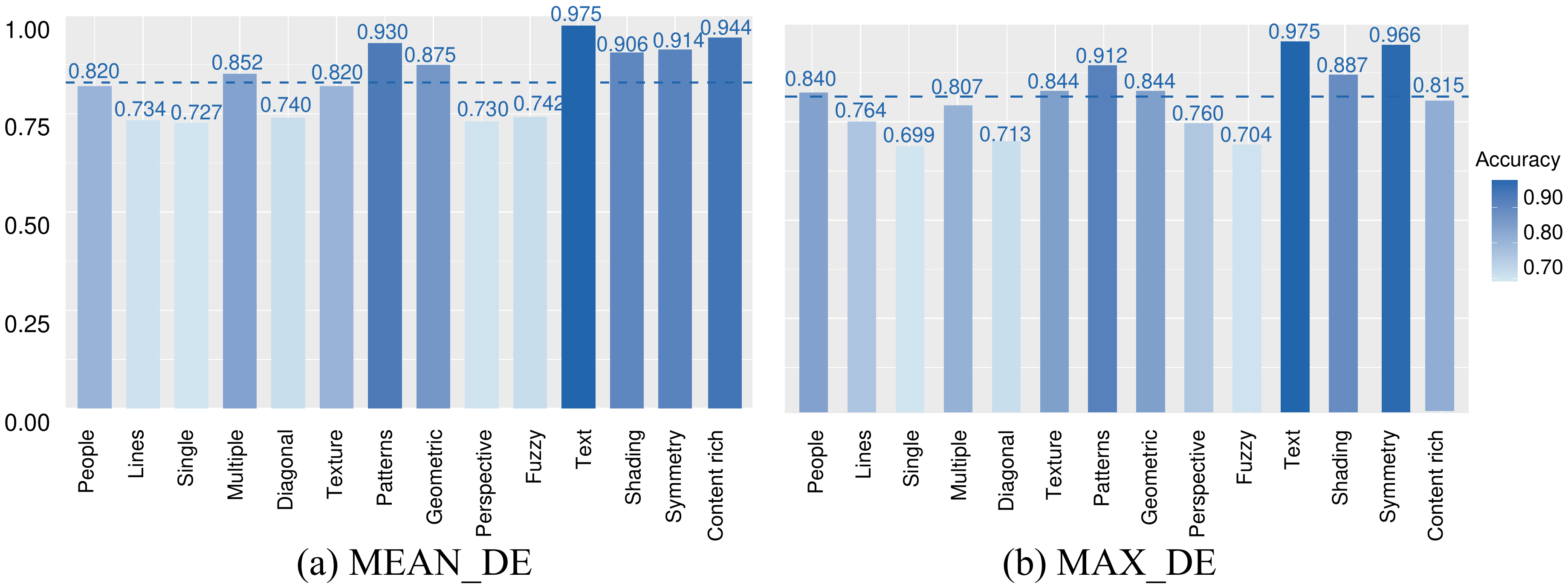}
  \caption{Accuracy rate for attribute prediction.}
  \label{fig:att_pre}
\end{figure}

In building the data set, we use the max rating (MAX-De) of the four methods from the six raters as the measurement of image retargetability.
One alternative method is to use the mean value, MEAN-De.
We trained the proposed siamese network together with $Net_{-}$ under the definition of MEAN-De.
RMSE of MEAN-De is as follows: ours$ = 0.27$ and $Net_{-} = 0.42$.
We label the testing samples as $1$ or $-1$ according to its ground truth retargetability: $1$ if the score is above than a threshold $\sigma$, and $-1$ otherwise.
Thereafter, the regression task can be evaluated as a binary classification task.
In Fig.~\ref{fig:roc}, we plot the receiver operating characteristic curve and report the area under curve (AUC) value by setting $\sigma$ to $0.95$ and $0.7$, respectively.
Due to the plot of ROC curve is based on the prediction of each test samples, we could not repeat this result by randomly splitting the data set five times.
For definition discussion, we randomly select $2000$ testing samples from one of the five testing processes.
Although the AUC value of MEAN-De is higher than that of MAX-De, RMSE of MAX-De is lower than that of MEAN-De.
Fig.~\ref{fig:att_pre} reports the results of the binary visual attribute learning.
Additional discussions about the definition are provided in the supplemental materials.

\section{Applications}
\label{sec:applications}

This section shows several applications of the proposed method and the dataset, including retargeting method selection, retargeting method assessment, and generating photo collage.
\begin{figure*}[t!]
\centering
 \includegraphics[width=0.99\linewidth]{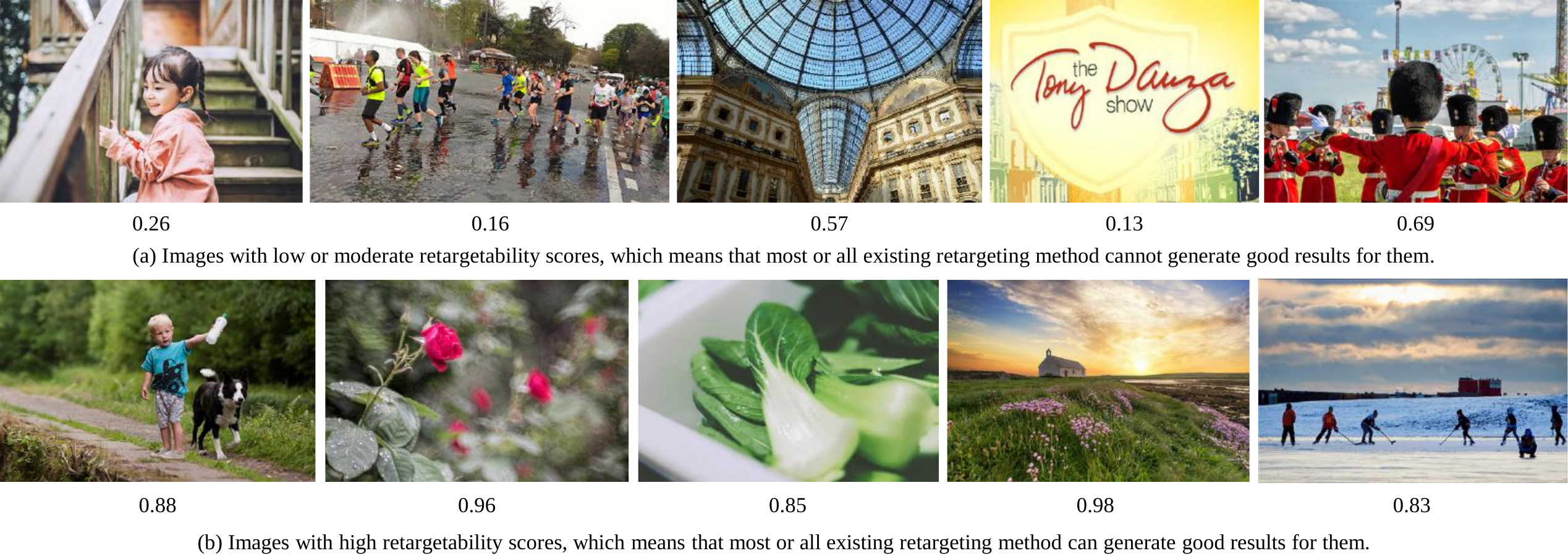}
  \caption{Images with different retargetability scores. The images in the first row are reliable for assessing new retargeting method.}
  \label{fig:app_img_sel02}
    \vspace{-2mm}
\end{figure*}

\begin{figure*}[tph!]
  \centering

 \subfigure[Input photos with various retargetability from low to high ]{
  \includegraphics[width=0.98\linewidth]{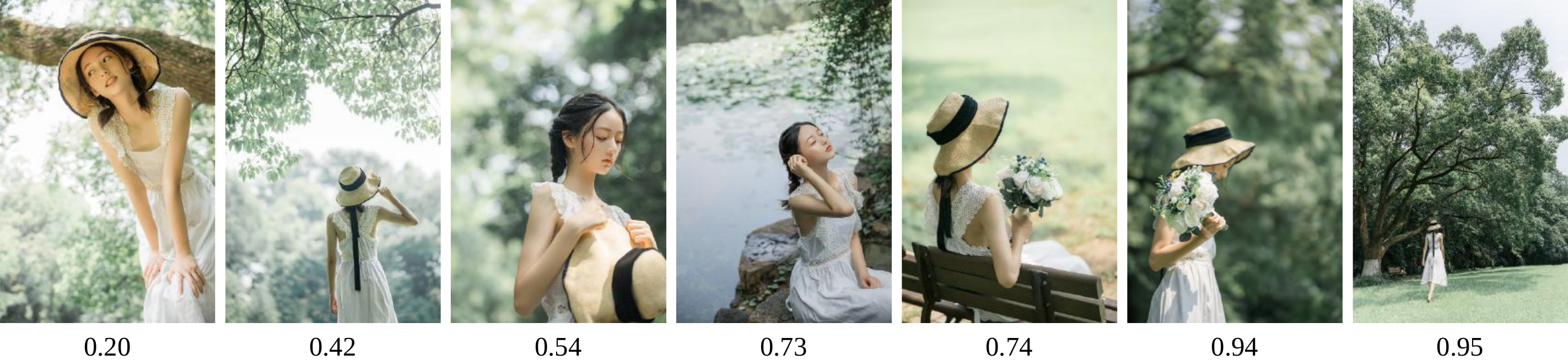}
  \label{fig:collage_set}
  }

  \subfigure[Our photo collage result by considering retargetability]{
  \includegraphics[width=0.48\linewidth]{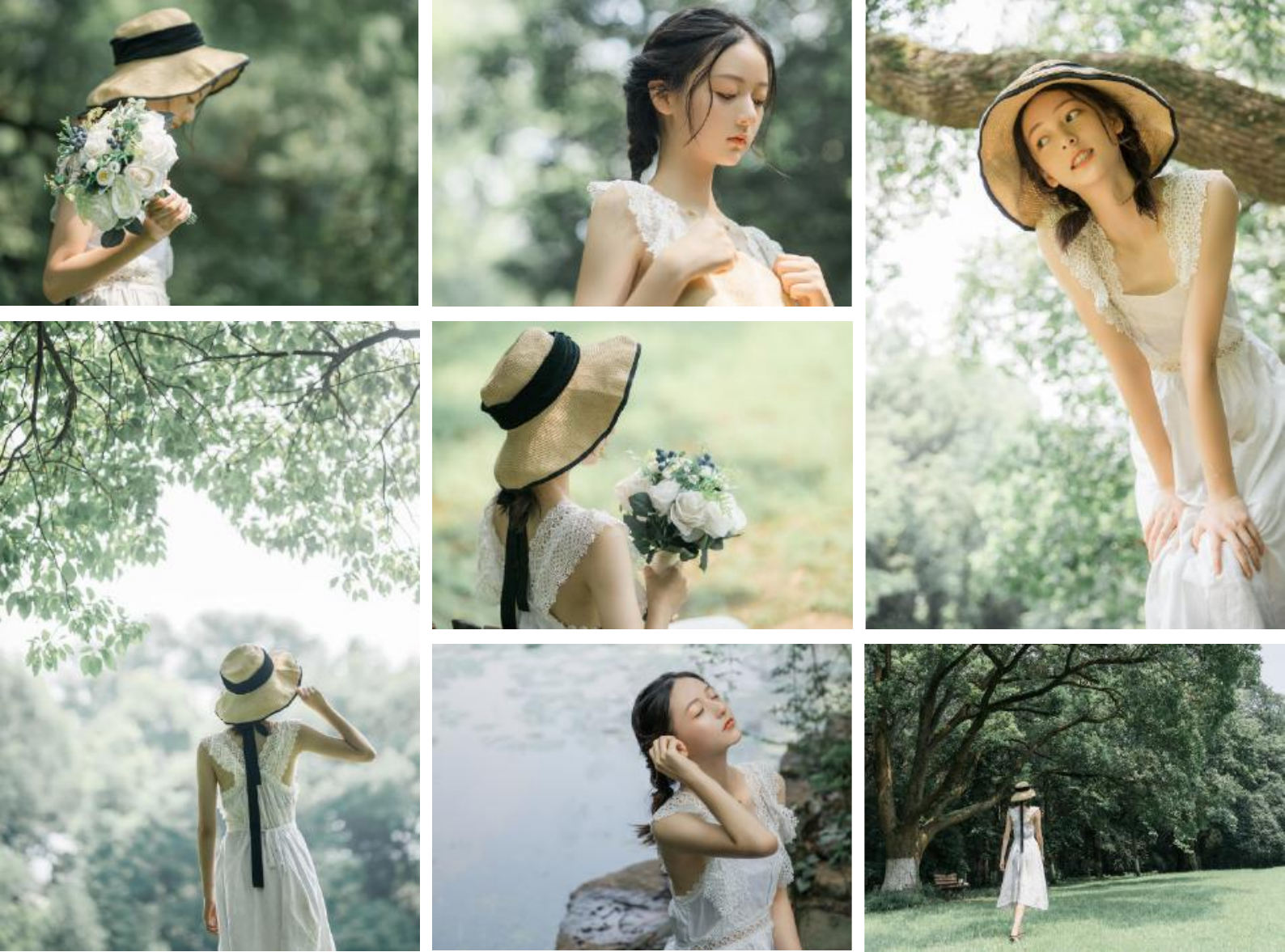}
  \label{fig:co_our}
  }
  \subfigure[Photo collage result without considering retargetability]{
  \includegraphics[width=0.48\linewidth]{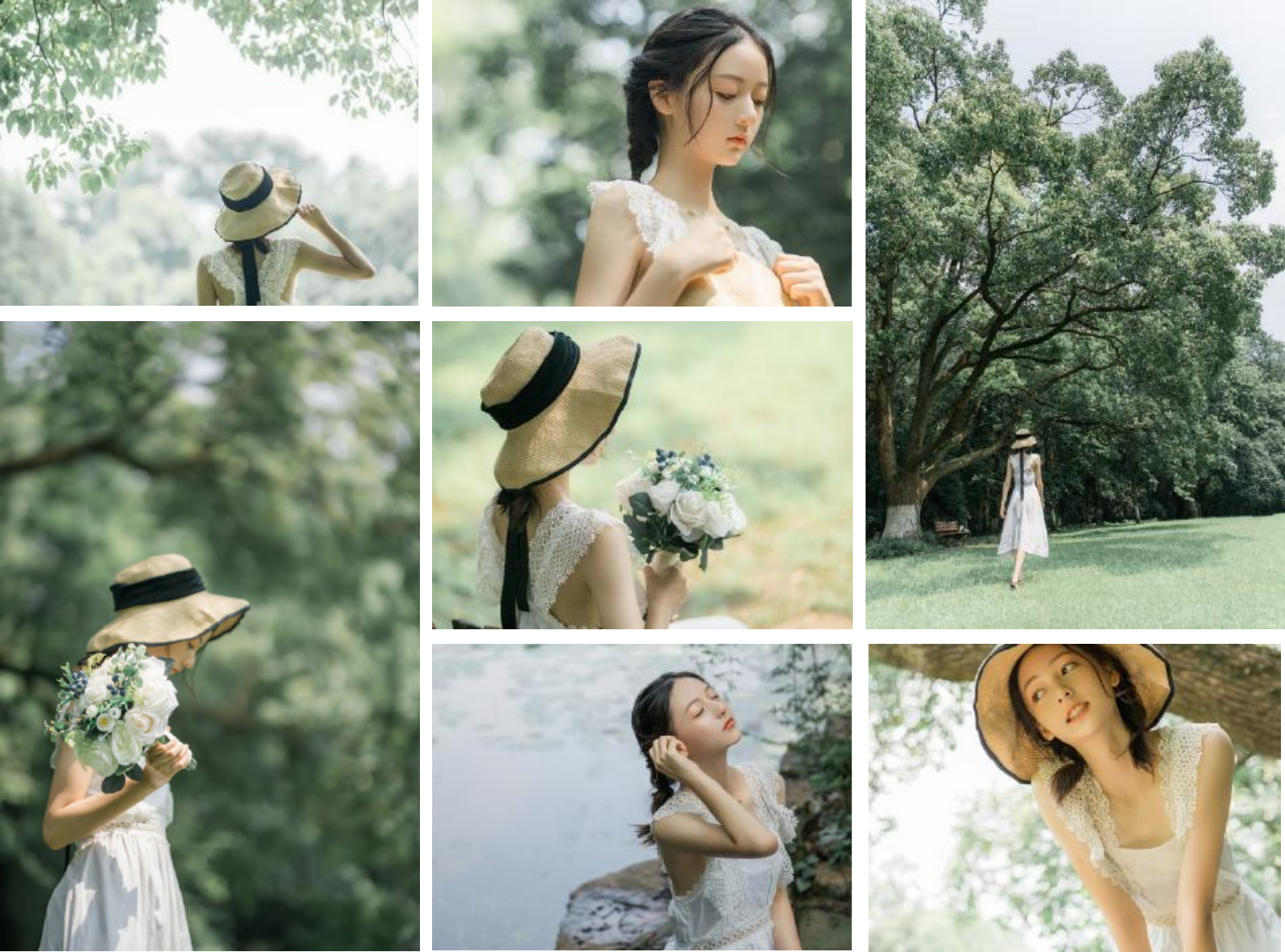}
  \label{fig:co_random}
  }

  \caption{Example of photo collage generation. We generated photo collage by using retargetability to guide the placement of photos. Our result preserved more salient content and presentd less retargeting artifact compared with the result without considering retargetability.}
  \label{fig:collage}
\end{figure*}

\subsection{Joint representations for retargeting tasks}

This study proposes a unified framework for joint learning visual attributes and image retargetability. 
Traditional CAIR (assessment) works highly depend on hand-crafted features. 
However, image retargetability is based on the features learned by a proposed end-to-end siamese network. 
The outputs of the last hidden full connected layer in the retargetability branch embed the image visual attributes and features related to the CAIR tasks. 
These joint representations offer insight into the possible connections between CAIR research and deep learning approaches.

We adopt the learned image representations for another CAIR task: \textit{retargeting method selection}, thereby suggesting the ``best'' retargeting method for a given image.
First, we collect images that have either ``good''  or ``acceptable'' retargeting results in the training set and record the ``best'' method(s) of each image based on manually annotations (Section~\ref{sec:dataset_methods}).
Second, we train the SVM classifier~\cite{CC:2001:libSVM} for each CAIR method to learn whether an image can be well retargeted by the method.
The inputs of these classifiers are the learned representations by our approach.
During testing, the method with the highest predicted value are suggested as the ``best'' retargeting method for a given image.
The average precision of the ``best'' retargeting method classification task is $82.12\%$ on the testing set.
Fig.~\ref{fig:qualitative} shows three such examples, in which the results of the suggested methods and results of other methods are compared.
We conduct a user study to evaluate the perceived quality from an observer's perspective.
In accordance with the experimental setting by \cite{Rubinstein:2010}, we adopt the paired comparisons technique, in which the participants are shown the original image and two retargeted images side by side.
One of retargeted images is the ``best'' result predicted by our classifiers and the other is randomly chosen from the four retargeted results.
The subjects are asked to compare the two results and choose the one they like better.
A third option called ``comparable'' is offered when participants find no marked difference between the two results.
A total of $300$ images from the testing set are selected for the user study.
During the survey, we set up a vigilance comparison every 10 tests, in which the two retargeted images are the same.
The results are discarded if one subject fails $50\%$ of the vigilance comparisons.
The vigilance comparisons ensure that workers are focusing, thereby leading to high-quality results.
A total of $82$ participants (age range of $20$-$45$) from different backgrounds are involved.
Among which, $95.12\%$ results are valid and we obtain $23,400$ votes.
Table~\ref{tab:userstudy} shows the statistics.
\begin{table}[]
\centering
\caption{Statistics for user study.}
\label{tab:userstudy}
\begin{tabular}{p{1.3cm}p{1.2cm}p{1.1cm}p{1.2cm}|p{1.3cm}}
\hline \hline
Option  &Adaptive Selection & Random Selection  & Comparable    & Vigilance       \\ \hline
Counts   &$13,164$           & $3,110$           & $4,786$       & $2,340$       \\ \hline
$\%$    &$56.26\%$          & $13.29\%$         & $20.45\%$     & $10.00\%$\\ \hline \hline
\end{tabular}
\end{table}
The ratio of the participants' selection between ``adaptive selection'' and ``random selection'' was $56.26\%:13.29\%\approx4.23$, which is consistent with that of the quantitative analysis ($82.12\%:17.88\%\approx4.59$).
The quantitative analysis and user study show that ``adaptive selection'' is superior to random guessing.
This finding is due to the fact that all the CAIR methods exhibit their own philosophies and each one worked better than the others for some images.
This result necessitates an adaptive selection for the ``best'' retargeting method for a given image example.

\subsection{Retargeting method assessment}

Although the CAIR methods have recently drawn considerable attention, the most popular assessment benchmark, namely, ``RetargetMe''~\cite{Rubinstein:2010}, was introduced approximately $10$ years ago.
The annotated dataset is relatively small and the current state-of-the-art CAIR methods report near-perfect results on this dataset. 
The current study offers a relatively large image dataset together with retargeting annotations, which can be used to augment other datasets, such as ``RetargetMe''~\cite{Rubinstein:2010}.
With the help of retargetability, people can easily collect a suitable testing set, which contains a wide range of images with different retargeting difficulties, to help the assessment of retargeting methods.
This scheme has also been used in image retrieval method evaluation by organizing the evaluation set to three different levels of difficulties~\cite{RO:2018:revisiting}.
To evaluate if a new proposed CAIR method is effective, the new method must be tested on images which are difficult to existing methods. 
Therefore, during assessment, we can just use images with low or moderate retargetability scores, such as the images in Figs.~\ref{fig:app_img_sel}(b), \ref{fig:app_img_sel}(c) and \ref{fig:app_img_sel02}(a).
People can use retargetability to filter some examples, which can be well retargeted by exiting methods such as the image in Figs.~\ref{fig:app_img_sel}(a) and \ref{fig:app_img_sel02}(b).
Through our experiments, we find that the images with retargetability arranging between $(0.0, 0.75]$ are reliable for the new CAIR method assessment.

\subsection{Photo collage generation}

Photo collage is often created by placing multiple photo images on a canvas of limited size.
The input images can be fitted on the canvas by retargeting them at the risk of losing important visual information and making the collage dull.
Therefore, optimally selecting image examples for different sizes of canvas regions is important because any input image cannot merely be well retargeted in a given scale.
Given the notion of image retargetability, automatic photo collage can be reliable
We present an example of using image retargetability to guide the generation of photo collage (see Fig.~\ref{fig:collage}).
The origin photos with varying retargetability are shown in Fig.~\ref{fig:collage}(a).
With the help of retargetability, the collage can be created in a simple but effective manner.
We first sort all the images based on their retargetability and place the images thereafter into the canvas in increasing retargetability order.
Images with relatively low retargetability are preferentially placed into regions where the aspect ratio can be retained to the maximum extent.
Fig.~\ref{fig:collage}(b) shows the result of our collage generation using this strategy.
Without considering retargetability, the collage may result in Fig.~\ref{fig:collage}(c), which causes severe content loss or boundary discontinuity artifacts to the photos (see the images in the left-top and right-bottom corners of Fig.~\ref{fig:collage}(c)).
We observe that considering retargetability can preserve salient content and present less unnatural retargeting artifact.

\section{Conclusion and Future Work}

This research presents retargetability as a novel image property and develops a computational predictor on the basis of multi-task learning. 
We construct a large image data set and annotate the retargetability of each image according to the quality of its retargeted results. 
We propose a siamese network structure that jointly learns attribute features and the relative retargetability. 
Our experiments show that image retargetability can be learned and predicted computationally and can be used to adaptively select a retargeting method for an image, find feasible image samples for retargeting method evaluation, and optimize collage layout for graphic design.
\begin{figure}
  \centering
  \includegraphics[width=0.95\linewidth]{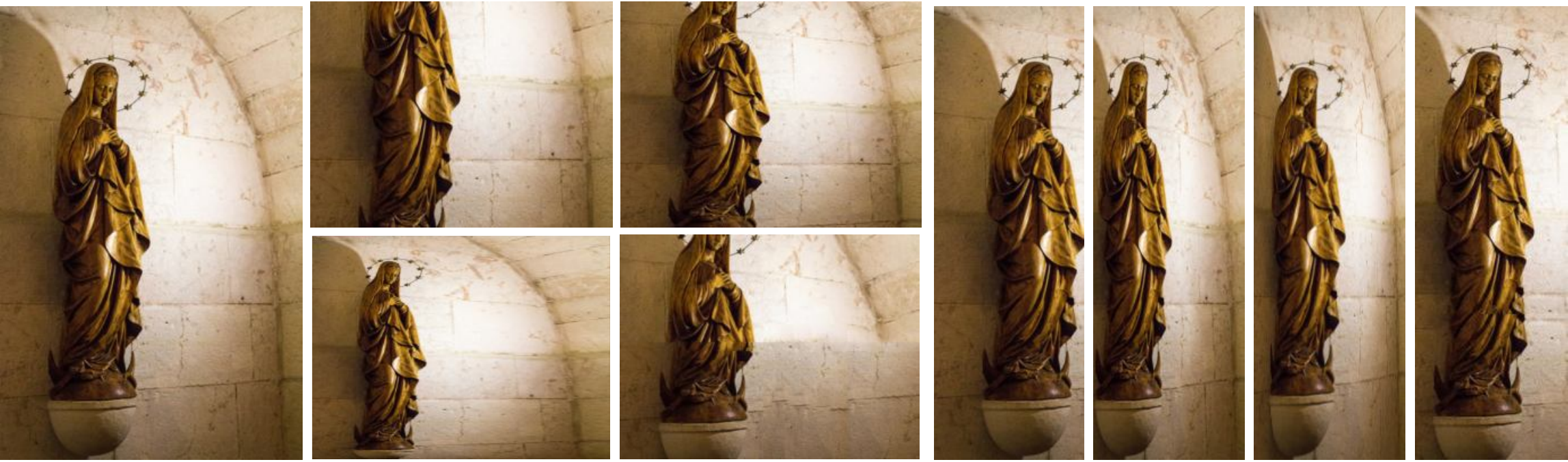}
  \caption{Retargeting along different dimensions. The input image is shown on the left, the retargeting results along the long and short dimension using the four selected CAIR methods are shown in the middle and right respectively.}
  \label{fig:limitation}
\end{figure}
\begin{figure}
\centering
\subfigure[Original Image ]{
  \includegraphics[width=0.34\linewidth]{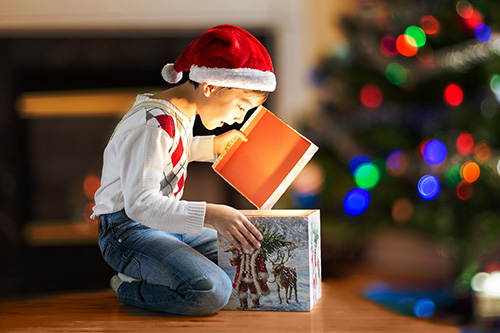}
  \label{fig:boy01}
  }
  \subfigure[$75\%$]{
  \includegraphics[width=0.255\linewidth]{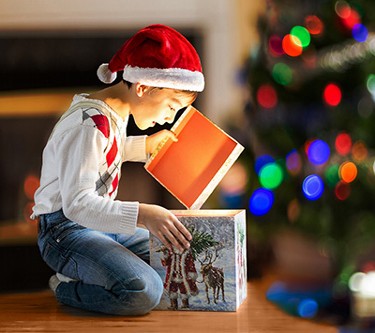}
  \label{fig:boy01_75}
  }
  \subfigure[$50\%$]{
  \includegraphics[width=0.17\linewidth]{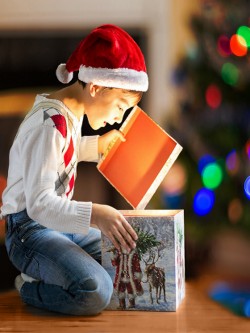}
  \label{fig:boy01_50}
  }
  \subfigure[$25\%$]{
  \includegraphics[width=0.085\linewidth]{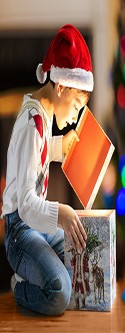}
  \label{fig:boy01_25}
  }
 \caption{Retarget one image to different scales. Targeting scale affects quality.}
  \label{fig:scale}
\end{figure}

Our experiments only consider the retargetability of an image in one dimension (i.e., long side), thereby indicating the we restrict the change to either the width or height of the image.
However, the retargetability of an image on these two dimensions may not consistently be similar. 
Fig.~\ref{fig:limitation} shows that when we retarget the long side, the resulting images may not be as satisfactory as retargeting the short side.
Therefore, we eventually plan to investigate the computation of image retargetability in both dimensions.
Another limitation of our method is that we only retargeted source images to a fixed scale ($50\%$), but the retargetability of an image may vary with the changing of target scale.
Fig.~\ref{fig:scale} shows that we retarget one image to $75\%$, $50\%$ and $25\%$ and we could see that the quality of the retargeting results is related to the targeting scale.

We can augment the resulting images and annotate them to analyze the relationship between retargetability and target size in the future.
Furthermore, we will attempt to generalize retargetability for analysis and processing of video data.

\section*{Acknowledgements}
We thank the anonymous reviewers for valuable comments.
This work was supported by National Key R\&D Program of China under no. 2018YFC0807500, and by National Natural Science Foundation of China under nos. 61832016, 61672520 and 61702488, and by Ministry of Science and Technology under no. 108-2221-E-006-038-MY3, Taiwan and by CASIA-Tencent Youtu joint research project.

\ifCLASSOPTIONcaptionsoff
  \newpage
\fi

\bibliographystyle{IEEEtran}
\bibliography{Retargetability}

\vspace{-15 mm}
\begin{IEEEbiography}[{\includegraphics[width=1in,height=1.25in,clip,keepaspectratio]{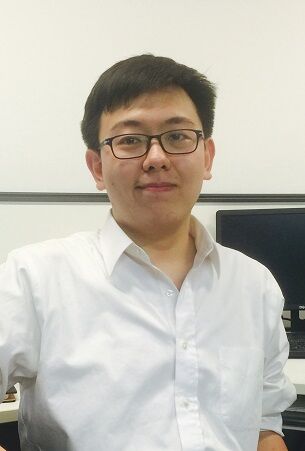}}]{Fan Tang} received the PhD degree in National Laboratory of Pattern Recognition, Institute of Automation, Chinese Academy of Sciences in 2019. He received the BSc degree in computer science from North China Electric Power University in 2013. He is currently a Post-Doctoral Scholar with Institute of Software, Chinese Academy of Sciences. His research interests include image synthesis and image recognition.
\end{IEEEbiography}
\vspace{-15 mm}
\begin{IEEEbiography}[{\includegraphics[width=1in, height=1.25in,clip,keepaspectratio]{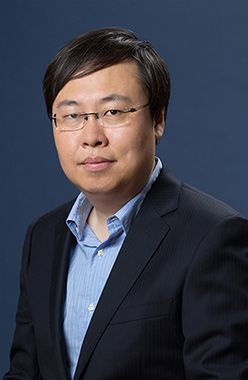}}]{Weiming Dong}
is a Professor in the Sino-European Lab in Computer Science, Automation and Applied Mathematics (LIAMA) and National Laboratory of Pattern Recognition (NLPR) at Institute of Automation, Chinese Academy of Sciences. He received the PhD in Computer Science from the University of Lorraine, France, in 2007. He received the BEng and MEng degrees in Computer Science in 2001 and 2004, both from Tsinghua University, China. His research interests include image synthesis and image recognition. Weiming Dong is a member of the ACM and IEEE.
\end{IEEEbiography}
\vspace{-13 mm}
\begin{IEEEbiography}[{\includegraphics[width=1in,height=1.25in,clip,keepaspectratio]{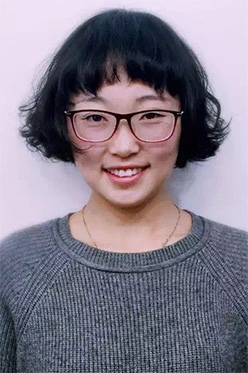}}]{Yiping Meng}
received the BSc degree in Software Engineering from University of Electronic Science and Technology of China in 2013. She received the MEng degree in National Laboratory of Pattern Recognition, Institute of Automation, Chinese of Sciences, in 2017. She is currently a research outreach manager in Didi Chuxing. Her research interests include image synthesis and image recognition.
\end{IEEEbiography}
\vspace{-13 mm}
\begin{IEEEbiography}[{\includegraphics[width=1in,height=1.25in,clip,keepaspectratio]{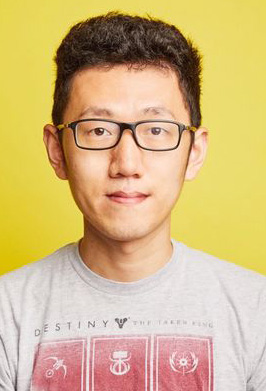}}]{Chongyang Ma}
received B.S. degree from the Fundamental Science Class (Mathematics and Physics) of Tsinghua University in 2007 and PhD degree in Computer Science from the Institute for Advanced Study of Tsinghua University in 2012. He is currently a Research Lead at Kuaishou Technology. His research interests include computer graphics and computer vision.
\end{IEEEbiography}
\vspace{-13 mm}
\begin{IEEEbiography}[{\includegraphics[width=1in,height=1.25in,clip,keepaspectratio]{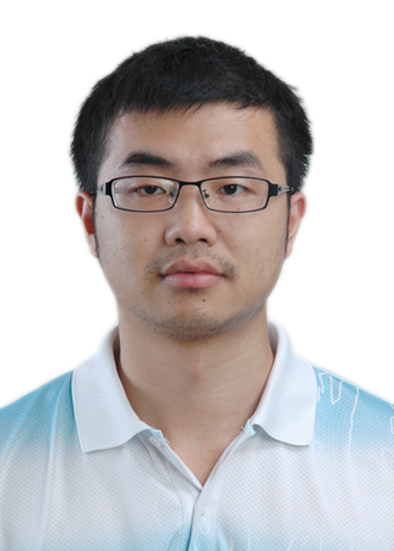}}]{Fuzhang Wu}
received the PhD candidate in the Sino-European Lab in Computer Science, Automation and Applied Mathematics (LIAMA) and National Laboratory of Pattern Recognition, Institute of Automation, Chinese Academy of Sciences. He received the BSc in computer science in 2010 from Northeast Normal University, P. R. China. His research interests include image synthesis and image analysis. He is currently a Post-Doctoral Scholar with Institute of Software, Chinese Academy of Sciences.
\end{IEEEbiography}
\vspace{-13 mm}
\begin{IEEEbiography}[{\includegraphics[width=1in,height=1.25in,clip,keepaspectratio]{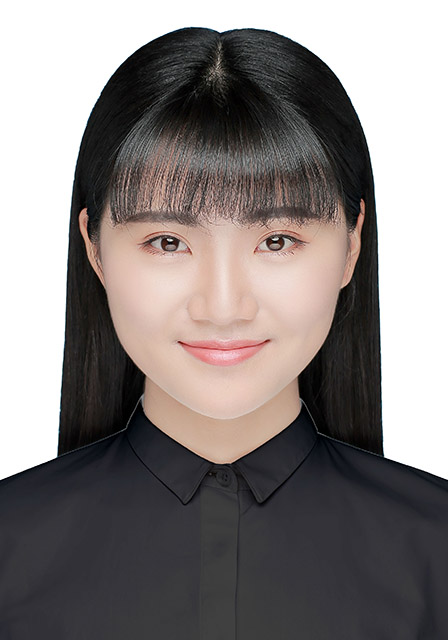}}]{Xinrui Li}
received the MSc degree from North China Electric Power University in 2019. She received the BEc degree in statistics from Shandong University of Finance and Economics in 2016. She is currently a researcher at State Power Investment Corporation Research Institute. Her research interests include applied statistics and mathematical statistics.
\end{IEEEbiography}
\vspace{-13 mm}
\begin{IEEEbiography}[{\includegraphics[width=1in,height=1.25in,clip,keepaspectratio]{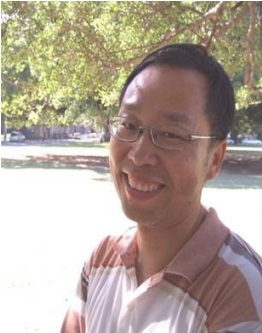}}]{Tong-Yee Lee}
received the PhD degree in computer engineering from Washington State University, Pullman, in May 1995. He is currently a chair professor in the Department of Computer Science and Information Engineering, National Cheng-Kung University, Tainan, Taiwan, ROC. He leads the Computer Graphics Group, Visual System Laboratory, National Cheng-Kung University (http://graphics.csie.ncku.edu.tw/). His current research interests include computer graphics, non-photorealistic rendering, medical visualization, virtual reality, and media resizing. He is a senior member of the IEEE and the member of the ACM.
\end{IEEEbiography}
\vspace{-13 mm}

\end{document}